\documentclass[10pt,twocolumn,letterpaper]{article}

\usepackage[pagenumbers]{cvpr}      

\usepackage{fix-cm}
\usepackage{float}
\usepackage{graphicx}
\usepackage{array}
\newlength{\CSUniformSize}
\newcommand{\CSUniformCell}[2]{%
  \begin{minipage}[t]{\linewidth}\centering
  \vspace{0pt}%
  \parbox[c][2.2\baselineskip][c]{\linewidth}{\centering\bfseries #1}\par\vspace{1pt}%
  \includegraphics[width=\CSUniformSize,height=\CSUniformSize]{#2}%
  \end{minipage}}
\newcommand{\CSPairedPrompts}[2]{%
  \noindent\begin{minipage}[t]{0.394\linewidth}%
  \vspace{0pt}\parbox[t][2.2\baselineskip][t]{\linewidth}{\raggedright\textit{Prompt:} #1}%
  \end{minipage}\hspace{0.014\linewidth}%
  \begin{minipage}[t]{0.592\linewidth}%
  \vspace{0pt}\parbox[t][2.2\baselineskip][t]{\linewidth}{\raggedright\textit{Prompt:} #2}%
  \end{minipage}\par\vspace{1pt}}

\definecolor{cvprblue}{rgb}{0.21,0.49,0.74}
\usepackage[pagebackref,breaklinks,colorlinks,allcolors=cvprblue]{hyperref}

\usepackage{microtype}
\usepackage{graphicx}
\usepackage{comment}
\usepackage{hyperref}
\usepackage[utf8]{inputenc}
\usepackage[T1]{fontenc}

\usepackage{xcolor}
\usepackage[most]{tcolorbox}
\usepackage{listings}
\usepackage{fancyvrb}
\usepackage{enumitem}
\usepackage{amsmath}
\usepackage{float}
\usepackage{booktabs}
\usepackage{graphicx}
\usepackage{multirow}
\usepackage{verbatim}
\usepackage{hyperref}

\usepackage{hyperref}
\usepackage{amsmath}
\usepackage{wrapfig} 
\usepackage{mathtools}
\usepackage{amsthm}
\usepackage{comment}
\usepackage{multirow}
\usepackage[table]{xcolor}
\usepackage[capitalize,noabbrev]{cleveref}

\usepackage{placeins}

\usepackage{pifont}
\newcommand{\cmark}{\ding{51}} 
\newcommand{\xmark}{\ding{55}} 
\newcommand{\circleone}{\ding{172}} 
\newcommand{\circletwo}{\ding{173}} 
\newcommand{\circlethree}{\ding{174}} 
\newcommand{\circlefour}{\ding{175}} 
\newcommand{\circlefive}{\ding{176}} 

\definecolor{promptbg}{rgb}{0.95, 0.95, 0.98}
\definecolor{correctnesscolor}{rgb}{0.0, 0.5, 0.0}
\definecolor{visualcolor}{rgb}{0.0, 0.0, 0.8}
\definecolor{efficiencycolor}{rgb}{0.8, 0.4, 0.0}
\definecolor{codegray}{rgb}{0.4, 0.4, 0.4}

\tcbuselibrary{breakable,skins}

\newtcolorbox{correctnessbox}[1]{
    colback=promptbg,
    colframe=correctnesscolor,
    boxrule=1pt,
    arc=3pt,
    left=8pt,
    right=8pt,
    top=8pt,
    bottom=8pt,
    breakable,
    title={\textbf{Correctness Evaluation Prompt}},
    fonttitle=\bfseries\large,
    #1 
}

\newtcolorbox{visualbox}[1]{
    colback=promptbg,
    colframe=visualcolor,
    boxrule=1pt,
    arc=3pt,
    left=8pt,
    right=8pt,
    top=8pt,
    bottom=8pt,
    breakable,
    title={\textbf{Visual Quality Evaluation Prompt}},
    fonttitle=\bfseries\large,
    #1
}

\newtcolorbox{efficiencybox}[1]{
    colback=promptbg,
    colframe=efficiencycolor,
    boxrule=1pt,
    arc=3pt,
    left=8pt,
    right=8pt,
    top=8pt,
    bottom=8pt,
    breakable,
    title={\textbf{Efficiency Evaluation Prompt}},
    fonttitle=\bfseries\large,
    #1
}

\newtcolorbox{renderbox}[1]{
    colback=promptbg,
    colframe=brown,
    boxrule=1pt,
    arc=3pt,
    left=8pt,
    right=8pt,
    top=8pt,
    bottom=8pt,
    breakable,
    title={\textbf{Prompt for Code Rendering}},
    fonttitle=\bfseries\large,
    #1
}

\newtcolorbox{understandingbox}[1]{
    colback=promptbg,
    colframe=gray,
    boxrule=1pt,
    arc=3pt,
    left=8pt,
    right=8pt,
    top=8pt,
    bottom=8pt,
    breakable,
    title={\textbf{Prompt for Understanding-level Planning}},
    fonttitle=\bfseries\large,
    #1
}

\def\confName{CVPR}
\def\confYear{2027}

\title{PlanViz: Evaluating Planning-Oriented Image Generation and Editing for Computer-Use Tasks}

\author{Junxian Li$^{1}$\thanks{These authors contributed equally.}, Kai Liu$^{1*}$, Leyang Chen$^{1*}$, Weida Wang$^2$, Zhixin Wang$^3$, \\Jiaqi Xu$^3$, Fan Li$^3$, Renjing Pei$^3$, Linghe Kong$^1$, Yulun Zhang$^1$\thanks{Corresponding to: yulun100@gmail.com} \\
\small{$^1$Shanghai Jiao Tong University, $^2$Fudan University, $^3$Huawei Technologies Ltd} \\
\texttt{\small{lijunxian0531@sjtu.edu.cn, yulun100@gmail.com}} 
}

\begin{document}
\maketitle
\begin{abstract}
  Unified multimodal models (UMMs) have shown impressive capabilities in complex natural image generation. However, their potential in computer-use planning tasks, which are closely related to our lives, remain rarely explored. Image generation and editing in computer-use tasks require capabilities like fine-grained perception and procedural understanding, and it is still unknown whether UMMs can finish these tasks well. Therefore, we propose PlanViz, a new benchmark designed to evaluate image generation and editing for computer-use tasks. We focus on sub-tasks which frequently involve in daily life and require planning steps. Specifically, three representative sub-tasks are designed: route planning, work diagramming, and web\&UI displaying. Data quality is ensured by including large human efforts on question, reference and criteria annotation. Additionally, a task-adaptive score, PlanScore, is proposed for comprehensive and exact evaluation. PlanScore helps understanding the correctness, visual quality and efficiency of generated images. Through experiments, we highlight key limitations and opportunities for future research on this topic. Codes and supplementary materials are in \url{https://github.com/lijunxian111/PlanViz}.
\end{abstract}
\vspace{-2mm}
\vspace{-2mm}
\section{Introduction}
\label{sec:intro}

Unified multimodal models (UMMs)~\cite{deng2025bagel, chen2025janus, openai2025b_gptimage1, wu2025omnigen2} have seen remarkable developments recently. They integrate the understanding and generation capabilities of multimodal large language models (MLLMs) and diffusion or flow models. Advanced UMMs can therefore overcome more complex tasks like creative drawing, multi-image fusion, and logical editing~\cite{pan2026wiseedit, chow2025weave, liang2026rover}. 
\begin{figure}[t]
    \centering
    \includegraphics[width=1\linewidth]{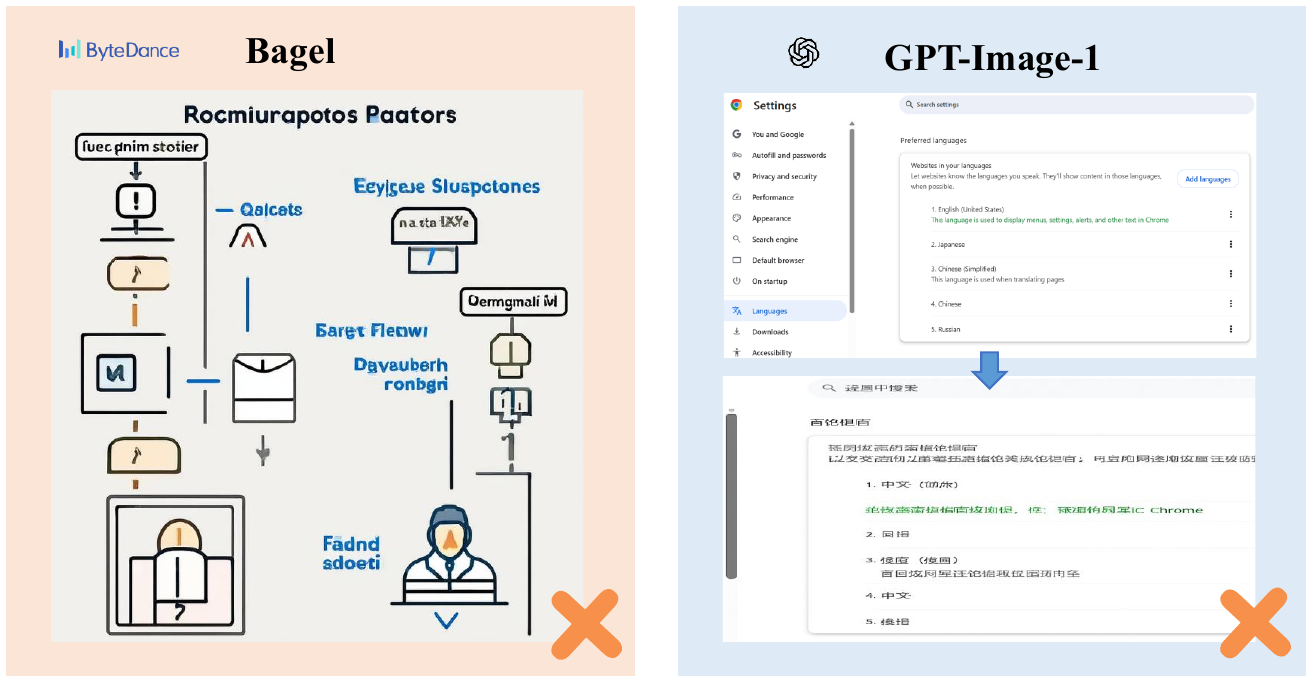}
    \caption{Examples of generation (left) and editing (right). The queries are ``generating a flowchart on how to apply a VISA'' and ``show what happens if setting `Chinese(Simplified)' to the display language''. Both UMMs make mistakes: Bagel doesn't provide a complete workflow and texts are meaningless; GPT-Image-1 provides garbled characters and fails to keep the total layout.}
    \label{fig:comp}
    \vspace{-6mm}
\end{figure}
Despite the huge success of UMMs in generating natural images, there exists a largely underexplored dimension: image generation and editing for computer-use tasks.  Unlike natural scenes, computer-related visual content, such as UI pages, structured workflows, and slide layouts, is practical to modern daily lives. Whether UMMs can accurately synthesize and manipulate such functional visuals for users directly reveal their utility as real-world assistants. Furthermore, these tasks demand a higher degree of structural precision and semantic consistency than typical image generation. Under these requirements, UMMs have to understand and precisely plan the tasks for generating high-quality images. Examples in Figure~\ref{fig:comp} support this perspective. Conversely, these requirements remain underexplored by current evaluation frameworks.

\begin{figure*}[t]
    \centering
    \includegraphics[width=\linewidth]{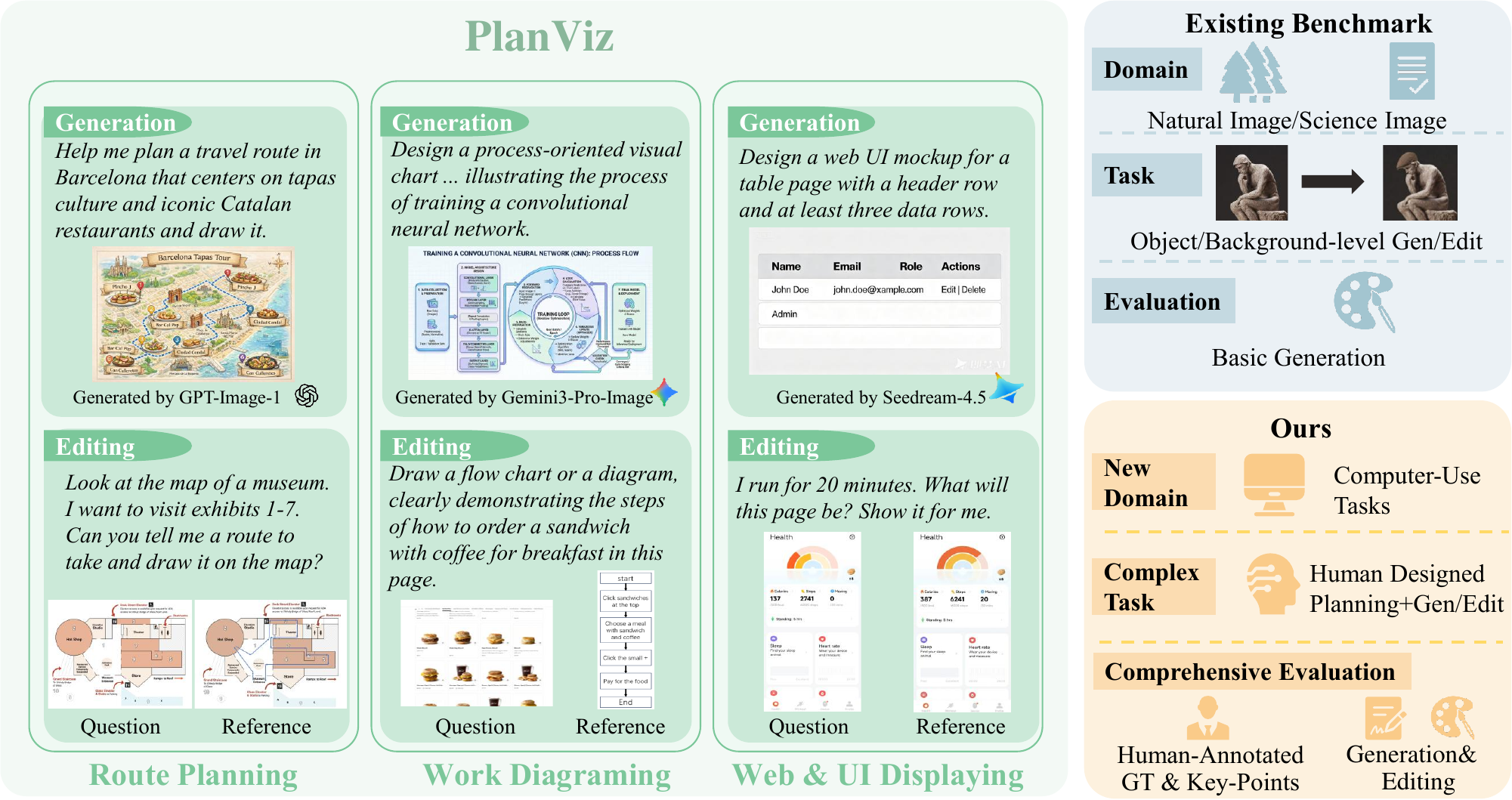}
    \vspace{-3mm}
    \caption{The overview of PlanViz. Our evaluation includes image generation and editing, with three proposed sub-tasks: route planning, work diagramming, and web\&UI displaying. Compared with prior benchmarks, PlanViz focuses on planning-oriented generation and editing in computer-use scenarios, supported by human-annotated references and keypoints.}
    \label{fig:intro_fig}
    \vspace{-6mm}
\end{figure*}

To fulfill this gap, we propose \textbf{PlanViz}, the first benchmark specially designed for image generation and editing in computer-use tasks with planning. Note that several challenges exist on constructing the benchmark: (i) specifying the target tasks for evaluation, (ii) collecting high-quality data, as existing datasets and benchmarks for UMMs mainly include natural images and instructions based on them, (iii) defining a detailed and exact judgment method since tasks may change a lot across images and instructions. 
To solve challenge (i), 
we consider two core aspects mentioned above across numerous computer-use tasks: \textbf{frequently involved in daily-life} and \textbf{including planning steps} and choose three kinds of sub-tasks: route planning, workflow diagramming, and web\&UI displaying, as shown in Figure~\ref{fig:intro_fig}.  To solve challenge (ii), we manually collect and annotate the whole dataset with quality control. Additionally, we get inspired from MLLM-as-judge and propose a new task-adaptive metric, PlanScore, for challenge (iii). PlanScore consists of three aspects: \textit{Cor}, \textit{Vis} and \textit{Ef}. These aspects measure the correctness and visual quality of generated images, together with whether they contain unnecessary or irrelevant content. Regarding real-life experience that sometimes tasks include exact destinations or goals and sometimes not, we also classify all questions into two categories: open-ended generation and closed-ended generation, to explore this effect. 

Extensive experiments are conducted on 12 UMMs and 14 models only for image generation and editing. 
Results illustrate that inconsistency on performance exists across almost all models and tasks. This is likely due to the mismatch between the models’ mostly reactive generation behavior and the explicit planning requirements imposed by different tasks. We also discover an obvious gap between open-source and closed-source proprietary UMMs. 
Models also tend to perform better on closed-ended questions than open-ended ones, suggesting that detailed, fine-grained guidance is important in our tasks. These findings highlight a direction for future research. In summary, our contributions are threefold:

\noindent $\bullet$ We propose PlanViz, a representative  benchmark relevant to real-life. To the best of our knowledge, this is the first benchmark evaluating image generation and editing in computer-use tasks with planning process. 

\noindent $\bullet$ We propose PlanScore, a task-adaptive judgment score including three aspects: correctness of generated images, visual quality, and measurements of unwanted parts. By utilizing MLLM-as-judge, this provides a detailed and human-supported evaluation process.

\noindent $\bullet$ We conduct a large-scale evaluation and analysis on 12 open-source or proprietary UMMs, together with 14 models only for image generation and editing. Results provide insightful findings about capabilities of UMMs in this new domain, and highlight areas for future improvement.
    
\section{Related Work}
\vspace{-1mm}
\noindent \textbf{Computer-use tasks.} 
Numerous sub-tasks are included like GUI interactions, workflow designing, map navigation and so on. Developing real-world AI assistants to solve these tasks has been a growing trend~\cite{cheng2024seeclick,sun2025seagent,sun2025gui,xu2026adamarp,dong2026neureasoner,dongaurora,jiang2026foe}. Therefore, we evaluate whether image generation or editing models, especially UMMs, can help finish these tasks well.

\noindent \textbf{UMMs.}
These models integrate understanding capabilities of MLLMs~\cite{grattafiori2024llama, bai2025qwen3, team2023gemini, zhu2025internvl3,liang2026pixelvla, zang2025reward,li2025chemvlm} and generation capabilities of diffusion or flow models~\cite{croitoru2023diffusion, rombach2022high, betker2023improving, blackforestlabs2024flux,wang2025spotactor,wang2024oneactor}. Recent works ~\cite{achiam2023gpt, wu2024vila,xie2024show, chen2025janus, deng2025bagel, li2025dual} try to develop end-to-end UMMs based on this idea, providing strong multimodal reasoning capabilities benefiting generation. The frontier proprietary models, GPT-Image-1~\cite{openai2025b_gptimage1} and Gemini3-Pro-Image~\cite{google2025nanopro} see a huge improvement in understanding very detailed and complex instructions and generating images.

\noindent \textbf{Previous benchmarks \& evaluation metrics for UMMs.}
Previous benchmarks typically contain only generating natural images. Some recent benchmarks include generating logically meaningful images or images requiring domain knowledge, like math or physics. UniEval~\cite{li2025unieval}, ROVER~\cite{liang2026rover}, WiseEdit~\cite{pan2026wiseedit} and GenExam~\cite{wang2025genexam} are some of them. 
However, they rarely explore whether UMMs can handle and plan various computer-use tasks and visualize them correctly. 
Similar to benchmarking image-generation models, evaluation metrics about image quality like LPIPS~\cite{zhang2018unreasonable}, FID~\cite{heusel2017gans} or object-level metrics~\cite{ghosh2023geneval} are frequently used. To evaluate whether generated images satisfy certain specific textual requirements, recent work~\cite{liang2026rover, wang2025genexam} has also considered using MLLM-as-judge~\cite{chen2024mllm}. Inspired by this, we design special evaluation metrics based on MLLM-as-judge for our task. 

\vspace{-2mm}
\section{PlanViz}
\vspace{-1mm}
\subsection{Benchmarking Objective}
\label{sec:motivation}

\begin{table}[t]
    \centering
    \setlength\tabcolsep{3pt}

    \caption{Comparison with other benchmarks. }
    \vskip -0.05in
    \resizebox{\linewidth}{!}{
    \begin{tabular}{l|lccc}
    \toprule
      \textbf{Benchmarks}  & Domain & Type & Planning & Keypoints  \\
    \midrule
    \rowcolor{blue!5} ROVER~\cite{liang2026rover}  & Natural/science image & Edit & \xmark &\xmark  \\
        GenExam~\cite{wang2025genexam} & Science(exam) image &  Gen & \xmark  &\cmark \\
     \rowcolor{blue!5}   WiseEdit~\cite{pan2026wiseedit} & Mostly natural image & Edit & \xmark  & \xmark \\
        Weave~\cite{chow2025weave} & Natural/science/Logic & Gen\&Edit & 
        partial & \cmark \\
     \rowcolor{cyan!10}   \textbf{PlanViz} (ours) & \textbf{Computer-use image} & \textbf{Gen\&Edit} & \textbf{\cmark} & \textbf{\cmark}  \\
    \bottomrule
    \end{tabular}
    }
    \vspace{-4mm}
    \label{tab:bench_comp}
\end{table}

PlanViz evaluates whether UMMs can turn goals and constraints in everyday computer-use scenarios into functional visual outcomes. We refer to this capability as planning-like visual generation: \textit{a model should understand the task requirements, organize the required spatial or procedural structure, and reflect the resulting plan in the generated image.}
We cover three representative task families: route planning, which emphasizes spatial constraints and destination ordering; workflow diagramming, which requires organizing multi-step procedures into structured visual representations; and web\&UI displaying, which tests whether models can infer and visualize interface layouts or states under user requirements and interactions. Extra explanations are in Appendix~\textcolor{cvprblue}{A}. Together, these tasks provide complementary views of planning-oriented generation and editing.

\noindent \textbf{Open-ended \& closed-ended. } Besides proposed sub-tasks, we notice that computer-use tasks in real-life sometimes include clear, detailed goals like buying a specified thing on a specified website, while sometimes only include rough directions like booking a room. Therefore, we categorize all questions as \textit{open-ended} and \textit{closed-ended}. Open-ended questions have no fixed answers, whereas closed-ended ones usually have more well-defined answers and allow fine-grained scoring. There are 231 open and 129 closed questions in total. Examples are in Appendix~\textcolor{cvprblue}{B}.

\begin{figure}[t]
    \centering
    \includegraphics[width=0.45\linewidth]{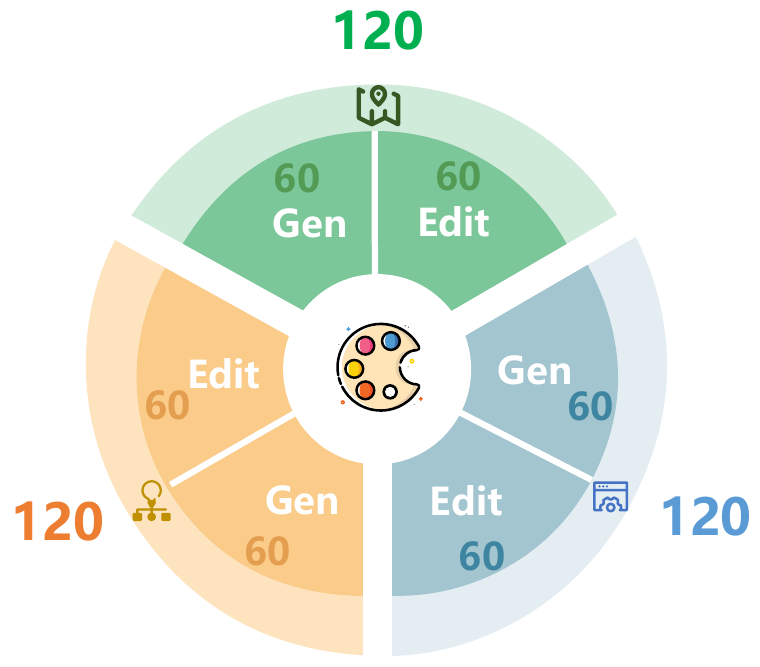} 
    \hfill
    \includegraphics[width=0.45\linewidth]{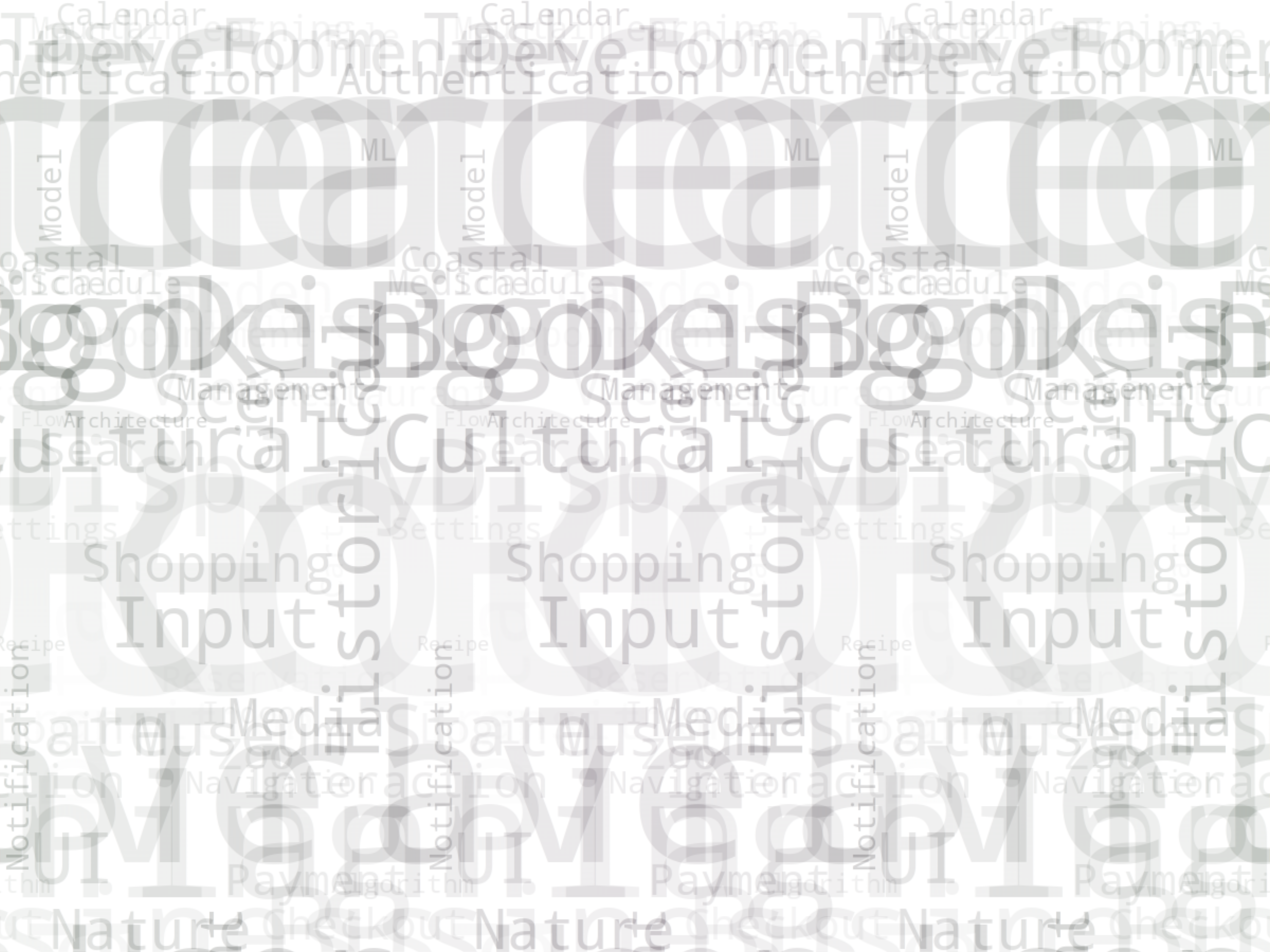} 
    \vspace{-2mm}
    \caption{The distribution (left) and the word cloud (right) of our benchmark. The green part represents route planning, while the orange part represents workflow diagramming, and the blue part represents web\&UI displaying. The word cloud shows the hot topics in all questions of our benchmark.}
    \label{fig:distribution}
    \vspace{-6mm}
\end{figure}

\begin{figure*}[t]
    \centering

    \includegraphics[width=\linewidth]{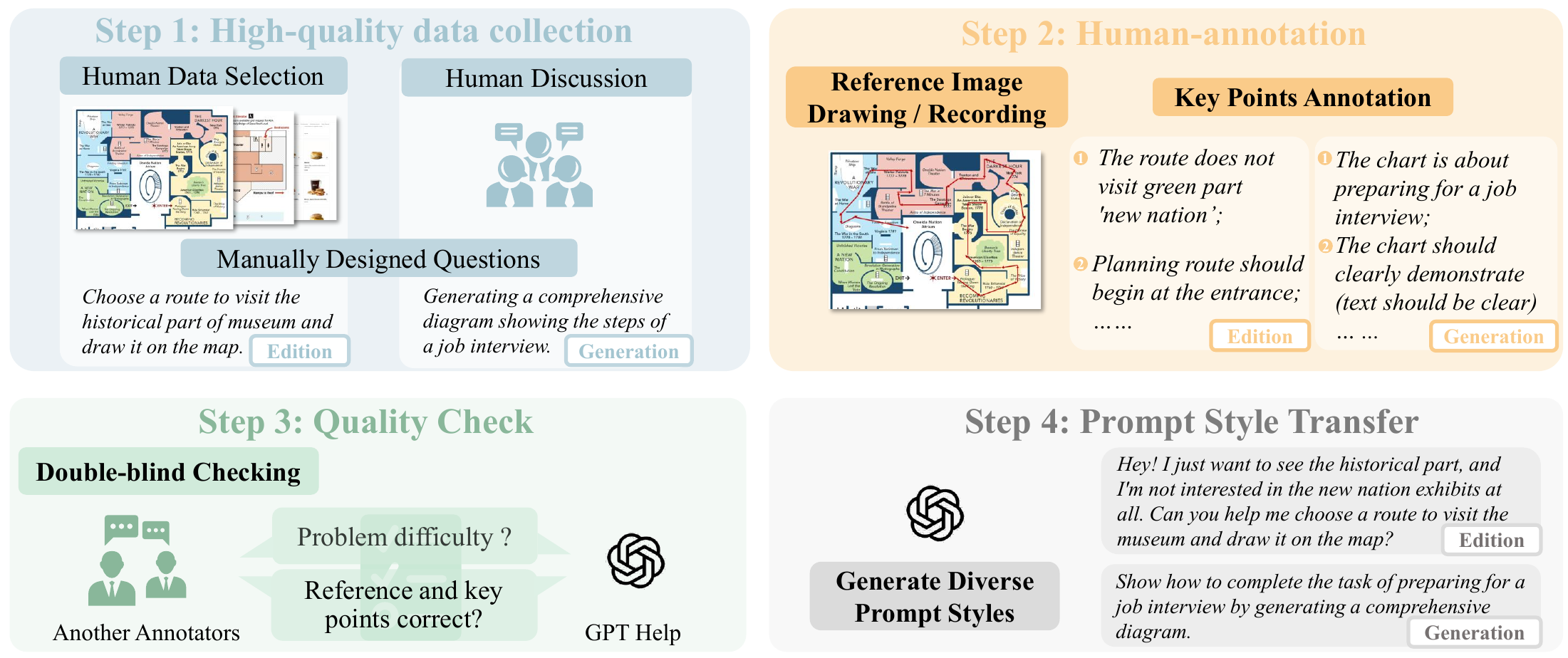}
    \caption{Pipeline of data construction. It consists of four stages: high-quality data collecting and cleaning, human annotation, quality check, and prompt style transformation. These stages are displayed from the top-left to the bottom right.}
    \label{fig:data_construct}
    \vspace{-4mm}
\end{figure*}

\subsection{Data Construction}
\label{sec:data_construct}
As illustrated in Figure~\ref{fig:data_construct}, we construct PlanViz
through a four-stage pipeline covering data collection, human annotation,
quality control, and prompt style transformation.

\textbf{Step 1. High-quality data collection.} For all \textbf{editing} sub-tasks, we manually\footnote{``manually'' here means that well-trained Ph.D. candidate students in Computer Science, who have led the construction of at least 2 other benchmarks, do this work manually. The same meaning is implied in the following context.} collect the source images from websites or apps. This step includes taking screenshots from various websites and apps. Detailed data sources and collection methods are justified in the Appendix~\textcolor{cvprblue}{D}.

We obey three key rules to collect high-quality images: (i) images should be clear, with the roads (route planning), patterns and layouts clearly visible, (ii) the texts, if they exist, should also be easy to recognize, (iii) chosen images are better to be taken from real-world maps, GUI screens, etc. Totally, nearly 500 images are collected. We then apply a careful manual scanning process on the collected data, details in Appendix~\textcolor{cvprblue}{C}.
Overall, the data distribution and topic-level word cloud are shown in Figure~\ref{fig:distribution}. The number of questions is similar to prior works~\cite{wang2025genexam, zhao2025envisioning,an2026genius}.

\textbf{Step 2. Human annotation.} 
Planning-oriented generation and editing may admit multiple valid visual
solutions. We therefore employ human annotators to construct
high-quality reference solutions and annotate task-specific keypoints
that describe the requirements each valid result should satisfy (Appendix~\textcolor{cvprblue}{E}). For
route planning, these keypoints may specify required destinations, path
constraints, or ordering requirements. For workflow diagramming, they
describe necessary steps and procedural relations. For web\&UI
displaying, they capture required interface elements, layouts, or state
changes. When applicable, annotators execute the requested action on
the real webpage or application and record the resulting screen as the
reference. These references and keypoints are later used in the
PlanScore evaluation pipeline.

\textbf{Step 3. Quality check.} We use a blind-test strategy here. In detail, another human annotators are recruited who do not know the ones build the dataset, to check two things\footnote{They do this by discussing together and asking GPT-5.1~\cite{openai2025gpt5}} whether: (i) the difficulty of each question is proper (ii) the annotated reference images and keypoints (Section~\ref{sec:score_judge}) are correct. If both are satisfied, this question is given 1 point; if not, 0 point is given. The total annotation score is 0.96, suggesting a strong agreement. For the controversial cases, all annotators are called together to modify them.

\textbf{Step 4. Prompt style transformation.} 
Finally, we diversify the linguistic styles of benchmark queries to better
reflect natural user expressions in real-world scenarios. GPT-5.1~\cite{openai2025gpt5} is used to paraphrase prompts at the surface level, including wording, tone, and
sentence structure, while preserving the original task semantics and
required planning information. We do not introduce any information that could provide prior advantages to closed-source models.

\subsection{Score Judgement Pipeline}
\label{sec:score_judge}

To tackle the problems in Section~\ref{sec:intro}, we propose a special task-adaptive score, \textbf{PlanScore}, to provide a detailed and accurate view of UMMs' abilities in this era.

Sample cases of generation results are checked to make sure the dimensions of our score (Appendix~\textcolor{cvprblue}{F}). Based on these observations, \textbf{PlanScore} consists of three dimensions. (i) Correctness (\textit{Cor}), which requires the models to generate \textit{correct} images successfully solving the task with \textit{obeying certain rules}. For instance, the planned route should follow existing roads, paths, etc.  (ii) Visual Quality (\textit{Vis}), which measures whether models generate \textit{visually and semantically coherent} images. For editing tasks in route planning and web\&UI displaying, the generated images are also evaluated to determine whether they keep the unchanged layout, details, objects, texts, etc. Notably, editing tasks in workflow diagramming require workflows as results, and these workflows do not need to be ``similar'' to the original websites. (iii) Efficiency (\textit{Ef}), which measures whether the models generate images without including items not required. 

\begin{figure}[t]
    \centering
    \includegraphics[width=\linewidth]{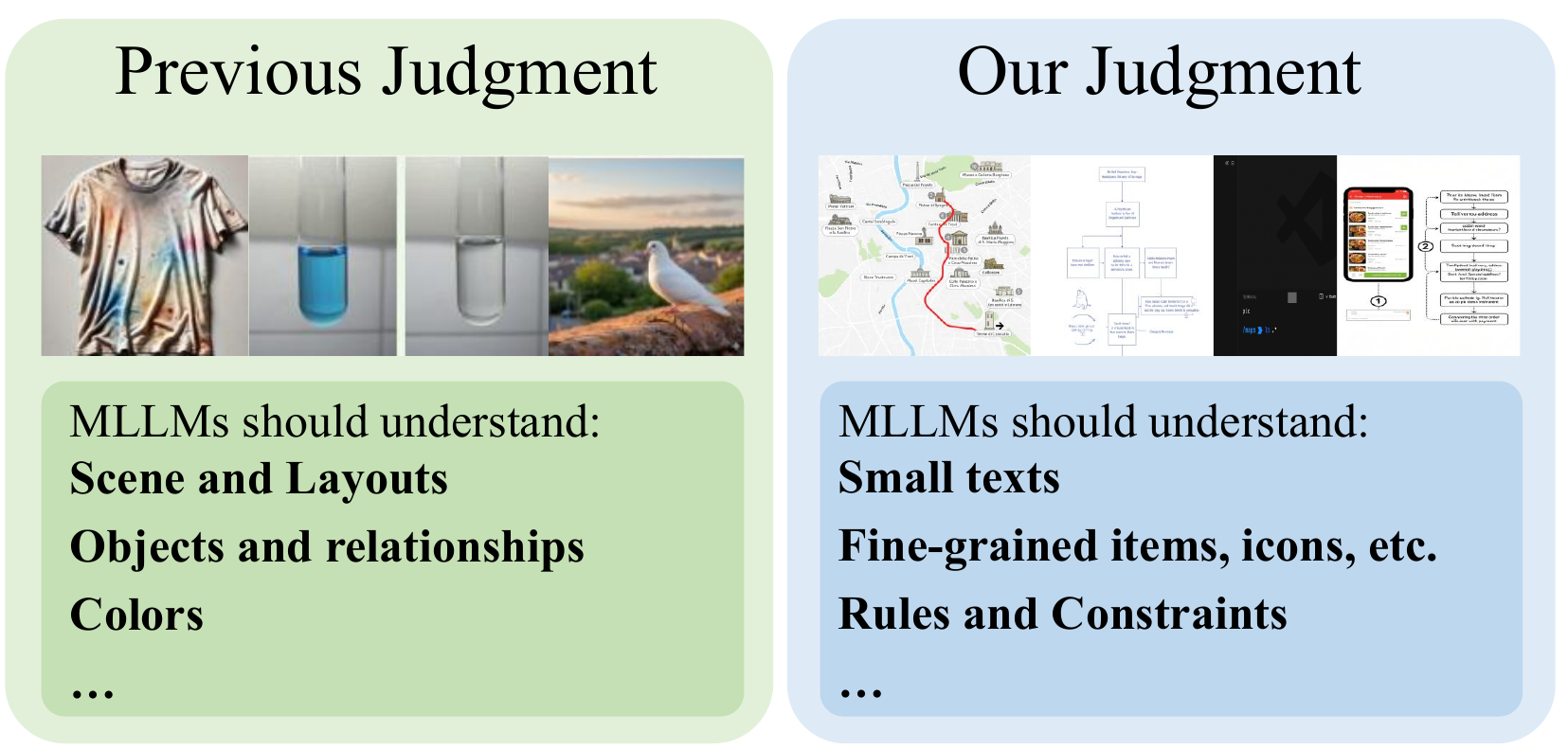}
    \caption{Comparison of previous judgments and ours.}
    \label{fig:eval_diff}
    \vspace{-5mm}
\end{figure}

We adopt a task-adaptive MLLM-as-judge
pipeline following prior works~\cite{wang2025genexam,liang2026rover,pan2026wiseedit}. As shown in
Figure~\ref{fig:eval_diff}, PlanViz requires fine-grained visual
understanding, small-text recognition, and reasoning over task-specific
rules and constraints. Balancing model capability and evaluation cost, we use Qwen3-VL-235B-A22B-Instruct~\cite{bai2025qwen3} as the judge model, with detailed scoring criteria provided for each
dimension. For \textit{Cor}, we introduce a key-point-based measurement. A unique set of key score points is annotated for each question. All scores are in $[0,1]$. Model details and evaluation prompts are included in Appendix~\textcolor{cvprblue}{G},~\textcolor{cvprblue}{H} and~\textcolor{cvprblue}{I}. As mentioned in Section~\ref{sec:data_construct}, human-annotated reference images are also given the MLLM, 
to help it better understand the tasks and what correct answers should be.  

\begin{table}[t]
    \centering
    \small
    \setlength\tabcolsep{3pt}
    \caption{Correlation between human and model judgments. }
    \vspace{-2mm}
    \begin{tabular}{c|ccc|ccc}
    \toprule
      \textbf{Models}  & \multicolumn{3}{c|}{OmniGen2} & \multicolumn{3}{c}{GPT-Image-1} \\
     \cmidrule{2-7}
     \textbf{\& Metrics} & \textit{Cor} & \textit{Vis} & \textit{Ef} & \textit{Cor} & \textit{Vis} & \textit{Ef} \\
      \midrule
      MLLM Mean & 0.20 & 0.71 & 0.52 & 0.65 & 0.84 & 0.86 \\
      Human Mean &  0.18  &   0.76  &  0.48    &   0.68   &  0.78    &   0.81   \\
    \midrule
    Human-MLLM \textit{r}  & 0.92 & 0.83 & 0.85 & 0.88 & 0.79 & 0.83 \\
    Human-MLLM MAE  & 0.02 & 0.05 & 0.04 & 0.03 & 0.08 & 0.05 \\
    \bottomrule
    \end{tabular}
    \label{tab:human}
\end{table}


\begin{table*}[t]
    \centering
    \caption{Main results on \textbf{image editing} tasks. ``Avg'' means the average of scores. All scores are between 0 and 1. We \textcolor{red}{\textbf{highlight}} the best performance of \textit{Cor}, for it is the most important score. We also add scores from our annotated reference images.}
    \vspace{-2mm}
    \setlength{\tabcolsep}{1.2mm}
    \resizebox{\textwidth}{!}{
    \begin{tabular}{l|r|cccc|cccc|cccc|c}
    \toprule
       \multirow{2}{*}{\textbf{Models}} &  \multirow{2}{*}{\textbf{Params}}  & \multicolumn{4}{c|}{\textbf{Route Planning}} & \multicolumn{4}{c|}{\textbf{Workflow Diagraming}} & \multicolumn{4}{c|}{\textbf{Web\&UI Displaying}} & \textbf{Overall} \\
    \cmidrule{3-15}
         &  & \textit{Cor} $\uparrow$ & \textit{Vis} $\uparrow$ & \textit{Ef} $\uparrow$ & Avg $\uparrow$ & \textit{Cor} $\uparrow$  & \textit{Vis} $\uparrow$ & \textit{Ef} $\uparrow$ & Avg $\uparrow$ & \textit{Cor} $\uparrow$  & \textit{Vis} $\uparrow$ & \textit{Ef} $\uparrow$ & Avg $\uparrow$ & Avg $\uparrow$ \\
         \midrule
        \rowcolor{cyan!10} \noalign{\renewcommand{\arraystretch}{0.5}} \multicolumn{15}{c}{Open-source Edit-only Models} \\
        \midrule
         AnyEdit~\cite{yu2025anyedit} & 1B & 0.00 & 0.82 & 0.70 & 0.51 & 0.00 & 0.50 & 0.31 & 0.27 & 0.01 & 0.56 & 0.52 & 0.36 & 0.36 \\
        \rowcolor{cyan!5} OmniGen~\cite{xiao2024omnigen} & 4B & 0.00 & 0.77 & 0.62 & 0.46 & 0.00  &  0.58 & 0.16 & 0.25 & 0.01 & 0.34 & 0.31 & 0.22 & 0.31 \\
         UltraEdit~\cite{zhao2024ultraedit} & 8B & 0.02 & 0.44 &  0.29 & 0.25 & 0.05 & 0.54 &  0.17 & 0.25 & 0.03 & 0.71 & 0.58 & 0.44 & 0.31 \\
\rowcolor{cyan!5} SD-3.5-large~\cite{rombach2022high} & 8B & 0.02 & 0.41 & 0.29 & 0.24 & 0.01   &  0.30 & 0.17 & 0.16 & 0.02 & 0.14 & 0.22 & 0.13 & 0.18 \\  
   Step1X-Edit-v1p2~\cite{liu2025step1x} & 12B & 0.02 & 0.75 & 0.64 & 0.47 & 0.03  & 0.49 & 0.29 & 0.27
   & 0.01  & 0.76 & 0.59 & 0.45 & 0.40 \\
    \rowcolor{cyan!5} Step1X-Edit-v1p2 (Thinking)~\cite{liu2025step1x} & 12B & 0.04 & 0.91 & 0.63 & 0.53 & 0.05  & 0.48 & 0.31 & 0.28 & 0.05 & 0.77 & 0.70 & 0.51 & 0.44 \\
      FLUX.1-Kontext-dev~\cite{blackforestlabs2024flux} & 12B & 0.11 & 0.57 & 0.42 & 0.37 & 0.02 & 0.36 & 0.20 & 0.19 & 0.00 & 0.57 & 0.40 & 0.32 & 0.29 \\
     \rowcolor{cyan!5}  Hidream-E1-Full~\cite{hidreami1technicalreport} & 17B &0.00 &0.33 &0.25 &0.19  &0.11 &0.19 &0.15 &0.15  &0.02  &0.26   &0.20  &0.16  &  0.17\\
      Qwen-Image-Edit~\cite{wu2025qwen} & 20B & 0.15 & 0.73 & 0.57 & 0.48 & 0.15 & 0.36 & 0.29 & 0.27 & 0.07 & 0.60 & 0.61 & 0.43 & 0.39 \\
    \midrule
        \rowcolor{lime!10} \noalign{\renewcommand{\arraystretch}{0.5}} \multicolumn{15}{c}{Open-source UMMs} \\
        \midrule
    Onecat~\cite{li2026onecat} & 3B & 0.00 & 0.16 & 0.33 &0.16  &0.00 &0.10  &0.12  &0.07  &0.02  & 0.09 &0.19  & 0.10 &0.11  \\
    \rowcolor{lime!5} Ovis-U1~\cite{wang2025ovisu1} & 3B & 0.17 & 0.32 & 0.19 & 0.23 & 0.08 & 0.31 & 0.13 & 0.17 & 0.05 & 0.41 & 0.30 & 0.25 & 0.22 \\
    Janus-4o~\cite{chen2025sharegpt} & 7B & 0.20 & 0.37 & 0.29 & 0.29 & 0.00 & 0.36 & 0.11 & 0.16 & 0.00 & 0.30 & 0.19 & 0.16 & 0.20 \\
    \rowcolor{lime!5} OmniGen2~\cite{wu2025omnigen2} & 7B & 0.27 & 0.84 & 0.78 & 0.63 & 0.00 & 0.69 & 0.44 & 0.38 & 0.05 & 0.69 & 0.56 & 0.43 & 0.48 \\
    UniPic2-Metaquery~\cite{wei2025skywork} & 9B & 0.26 & 0.72 & 0.52 & 0.50 & 0.06  & 0.46 & 0.13 & 0.22 & 0.04 & 0.37 & 0.40 & 0.27 & 0.33 \\
   \rowcolor{lime!5} UniPic2-Metaquery-GRPO~\cite{wei2025skywork} & 9B & 0.00 & 0.79 & 0.54 & 0.44 & 0.12 & 0.44 & 0.24 & 0.27 & 0.14 & 0.62 & 0.56 & 0.44  & 0.38 \\
     NextStep-1-Large-Edit~\cite{nextstepteam2025nextstep1} & 14B & 0.00 & 0.34 &    0.21 & 0.18 & 0.03 & 0.26 & 0.14 & 0.14 & 0.00 & 0.16 & 0.00 & 0.05 & 0.13 \\
   \rowcolor{lime!5} Bagel~\cite{deng2025bagel} & 14B & 0.04 & 0.03 & 0.73 & 0.27 & 0.05  & 0.33 & 0.05 & 0.14 &  0.07 & 0.02 & 0.06 & 0.05 & 0.15 \\
    Bagel (Thinking)~\cite{deng2025bagel} & 14B & 0.05 & 0.37 & 0.23 & 0.22 & 0.10  & 0.57 & 0.14 & 0.27 &  0.04 & 0.04 & 0.10 & 0.06 & 0.18 \\
    \midrule
        \rowcolor{orange!10} \noalign{\renewcommand{\arraystretch}{0.5}} \multicolumn{15}{c}{Closed-source proprietary Models} \\
        \midrule
    Wan-2.5-i2i-preview~\cite{wan2025wan} & --- & 0.12 & 0.70 & 0.56 & 0.46 & 0.18 & 0.57 & 0.41 & 0.39 & 0.18 & 0.82 & 0.78 & 0.59 & 0.48 \\
   \rowcolor{orange!5} Seedream-4.5~\cite{seedream2025seedream} & --- & 0.39 & 0.85 & 0.69 & 0.64 & \textcolor{red}{\textbf{0.51}} & 0.87 & 0.83 & 0.74 & \textcolor{red}{\textbf{0.30}} & 0.79 & 0.78 & 0.62 & 0.67 \\
   GPT-Image-1~\cite{openai2025b_gptimage1} & --- & \textcolor{red}{\textbf{0.42}} & 0.59 & 0.63 & 0.55 & 0.42 & 0.93 & 0.92 & 0.76 &  \textbf{\textcolor{red}{0.30}} & 0.57 &  0.75 & 0.53 & 0.61 \\
  \rowcolor{orange!5} Gemini3-Pro-Image~\cite{google2025nanopro} & --- &0.11 &0.06&0.53 &0.23&0.35 &0.01 &0.72  &0.36  & 0.12 &0.02   & 0.30  &0.15  & 0.25 \\
   \midrule
   Human Reference & --- & 1.00 & 0.96 & 0.98 & 0.98 & 1.00 & 0.97 & 1.00 & 0.99 & 1.00 & 0.98  & 0.97 & 0.98 & 0.98 \\
    \bottomrule
    \end{tabular}
    } 
    \label{tab:main_edit}
    \vspace{-3mm}
\end{table*}

\subsection{Evaluation Reliability}
\label{sec:human_study}
We first include a human study to test whether our judgment method is consistent with human intuition. Regarding~\cite{wang2025genexam,zhao2025envisioning}, 10 human annotators are recruited here. 30\% samples in our dataset are selected, individually from the generation results of OmniGen2 and GPT-Image-1 of both generating and editing tasks, and for each image generated by them, we employ our annotators to judge the PlanScore. we report the mean, Pearson correlation coefficient (\textit{r}) and Mean Absolute Error (MAE) in Table~\ref{tab:human}. This evaluation is consistent with prior works like~\cite{liang2026rover,wang2025genexam,zhao2025envisioning}. During evaluation, we notice that the MLLM sometimes assess higher \textit{Vis} and \textit{Ef} scores in our tasks, but not much. Study results suggest that a strong agreement exists between human judgments and MLLM's judgments with \textit{r}$>$0.8 mostly, MAEs in an acceptable scope. 

For better supporting judgment reliability, we employ GPT-5~\cite{openai2025gpt5} on these sampled data for cross-model evaluation (Appendix~\textcolor{cvprblue}{J}). This evaluation also suggests that our judge can reliably reflect the performances on PlanViz.

\begin{table*}[t]
    \centering
    \caption{Main results on \textbf{image generation} tasks. ``Avg'' means the average of scores. All scores are between 0 and 1. We \textcolor{red}{\textbf{highlight}} the best performance of \textit{Cor}, for it is the most important score on these tasks. ``Generation'' means image generation here.}
    \resizebox{\textwidth}{!}{
    \begin{tabular}{l|r|cccc|cccc|cccc|c}
    \toprule
       \multirow{2}{*}{\textbf{Models}} & \multirow{2}{*}{\textbf{Params}}  & \multicolumn{4}{c|}{\textbf{Route Planning}} & \multicolumn{4}{c|}{\textbf{Workflow Diagramming}} & \multicolumn{4}{c|}{\textbf{Web \& UI Displaying}} & \textbf{Overall} \\
    \cmidrule{3-15}
       &  & \textit{Cor} $\uparrow$ & \textit{Vis} $\uparrow$ & \textit{Ef} $\uparrow$ & Avg $\uparrow$ & \textit{Cor} $\uparrow$  & \textit{Vis} $\uparrow$ & \textit{Ef} $\uparrow$ & Avg $\uparrow$ & \textit{Cor} $\uparrow$  & \textit{Vis} $\uparrow$ & \textit{Ef} $\uparrow$ & Avg $\uparrow$ & Avg $\uparrow$ \\
       \midrule
        \rowcolor{cyan!10} \noalign{\renewcommand{\arraystretch}{0.5}} \multicolumn{15}{c}{Open-source Generation-only Models} \\
        \midrule
      OmniGen~\cite{xiao2024omnigen} & 4B & 0.01 & 0.85 & 0.75 & 0.54 & 0.01 & 0.43 & 0.20 & 0.21 & 0.21 & 0.75 & 0.24 & 0.40 & 0.38 \\
      \rowcolor{cyan!5} SD-3.5-large~\cite{rombach2022high} & 8B & 0.24 & 0.81 & 0.60 & 0.55 & 0.26  & 0.45 & 0.31 & 0.34 & 0.67  & 0.58 & 0.45 & 0.57 & 0.49 \\
      FLUX.1 dev~\cite{blackforestlabs2024flux} & 12B & 0.10 & 0.95 & 0.51 & 0.52 &  0.05 & 0.46  & 0.18 & 0.23 & 0.70  & 0.76 & 0.60 & 0.69 & 0.48 \\
      \rowcolor{cyan!5} Hidream-I1-Full~\cite{hidreami1technicalreport} & 17B & 0.03&0.85 &0.27 &0.38 &0.00   &0.56  &0.11  &0.22 & 0.06 & 0.54 &0.12 &0.24 & 0.28 \\
        Qwen-Image~\cite{wu2025qwen} & 20B & 0.46 & 0.74 & 0.66 & 0.62 & 0.77 & 0.71 & 0.84 & 0.77 & 0.77 & 0.84 & 0.78 & 0.80 & 0.73 \\
    \midrule
        \rowcolor{lime!10} \noalign{\renewcommand{\arraystretch}{0.5}} \multicolumn{15}{c}{Open-source UMMs} \\
        \midrule
    Onecat~\cite{li2026onecat} & 3B &0.04 &0.44  &0.50  &0.33  & 0.01& 0.22 &0.27  &0.17  &0.19  &0.30  &0.33  &0.27  &0.26  \\
      \rowcolor{lime!5}  Ovis-U1~\cite{wang2025ovisu1} & 3B & 0.11 & 0.79 & 0.66 & 0.52 & 0.15 & 0.43 & 0.19 & 0.24 & 0.27 & 0.48 & 0.35 & 0.37 & 0.38 \\
    Janus-4o~\cite{chen2025sharegpt} & 7B & 0.09 & 0.93 & 0.87 & 0.63 & 0.25 & 0.47 &  0.29 & 0.31 & 0.43 & 0.57 & 0.62 & 0.54 & 0.49 \\
    \rowcolor{lime!5} OmniGen2~\cite{wu2025omnigen2} & 7B & 0.13 & 0.88 & 0.65 & 0.55 & 0.23 & 0.43 & 0.23 & 0.22 & 0.60 & 0.71 & 0.58 & 0.63 & 0.47 \\
    UniPic2-Metaquery~\cite{wei2025skywork} & 9B & 0.02 & 0.87 & 0.45 & 0.45 & 0.00 & 0.55 & 0.05 & 0.20 & 0.50 & 0.60 & 0.56 & 0.55 & 0.40 \\
    \rowcolor{lime!5} UniPic2-Metaquery-GRPO~\cite{wei2025skywork} & 9B & 0.14 & 0.79 & 0.54 & 0.49 & 0.13  & 0.44 & 0.24 & 0.27 & 0.54  & 0.62 & 0.56 & 0.57 & 0.44 \\
     NextStep-1-Large~\cite{nextstepteam2025nextstep1} & 14B & 0.00 & 0.59 & 0.02 & 0.20 & 0.00 & 0.58 & 0.00 & 0.19 & 0.00  & 0.96 &  0.00 & 0.32 & 0.24 \\
    \rowcolor{lime!5} Bagel~\cite{deng2025bagel} & 14B & 0.07 & 0.94 & 0.73 & 0.58 & 0.09 & 0.39 & 0.13 & 0.20 & 0.35  & 0.62 & 0.44 & 0.47 & 0.42  \\
    Bagel-Thinking~\cite{deng2025bagel} & 14B & 0.42 & 0.85 & 0.70 & 0.66 & 0.22  & 0.56 & 0.39 & 0.39 & 0.55  & 0.69 & 0.69 & 0.64 & 0.56 \\
    \midrule
        \rowcolor{orange!10} \noalign{\renewcommand{\arraystretch}{0.5}} \multicolumn{15}{c}{Closed-source proprietary Models} \\
        \midrule
    Wan-2.5-t2i-preview~\cite{wan2025wan} & --- & 0.73 & 0.74 & 0.90 & 0.79 & \textbf{\textcolor{red}{0.91}} & 0.86 &  0.90 & 0.89 &  0.80 & 0.79 & 0.85 & 0.81 & 0.83 \\
   \rowcolor{orange!5} Seedream-4.5~\cite{seedream2025seedream} & --- & \textbf{\textcolor{red}{0.81}} &  0.78 & 0.89 & 0.83 & 0.88 & 0.94 & 0.92 & 0.91 & 0.86 & 0.92 & 0.88 & 0.89 & 0.88 \\
   GPT-Image-1~\cite{openai2025b_gptimage1} & --- & 0.67 & 0.96 & 0.98 & 0.87 & 0.89 & 0.97 & 0.92 & 0.93 & \textbf{\textcolor{red}{0.92}} & 0.92 & 0.90 & 0.91 & 0.80 \\
   \rowcolor{orange!5} Gemini3-Pro-Image~\cite{google2025nanopro} & --- & 0.78 & 0.89 & 0.80 & 0.82  & 0.89 & 0.82 &0.87 & 0.86 & \textbf{\textcolor{red}{0.92}} & 0.90  & 0.80  & 0.87  & 0.85 \\
    \bottomrule
    \end{tabular}
    }
    \label{tab:main_gen}
   \vspace{-3mm}
\end{table*}

\section{Experiment}
\label{sec:experiment}
\vspace{-1mm}
Recent UMMs, together with prior image generation and editing models, are evaluated. Main findings and cases are in Section~\ref{sec:main_results} and~\ref{sec:case_study}. 
Notably, \textit{Cor} is the most important score to measure whether correct and useful images are generated.

\subsection{Implementation}
\label{sec:implementation}
We apply recent state-of-the-art open-source or closed-source SOTA UMMs. Some prior models are also utilized. We evaluate all closed-source models through their official API. The list of models and evaluation details are in Appendix~\textcolor{cvprblue}{K} and~\textcolor{cvprblue}{L}. Notably, we randomly select a subset of 50 images from all generations and utilize the Python API of Qwen3-VL-235B-A22B-Instruct, mentioned in Section~\ref{sec:score_judge} to judge them 10 times. We discover that very similar scores are given, and after calculating the \textit{Coefficient of Variation}~\cite{abdi2010coefficient} ($\frac{\text{mean}}{\text{standard deviation}}\times100\%$), we gain a 3.4\% score, suggesting the stability of our evaluation.

\subsection{Benchmarking Results and Findings}
\label{sec:main_results}

\label{sec:findings}
Table~\ref{tab:main_edit} (editing) and Table~\ref{tab:main_gen} (generation) show the main results, highlighting five key findings: 

\circleone \textbf{ Challenges and strengths of PlanViz.} Image editing remains challenging with even SOTA models (0.61-0.67), trailing human-annotated ground truth by a wide margin. Additionally, image generation exhibits a sharp performance divide between closed-source proprietary models (e.g., Seedream 4.5, 0.88) and open-source ones (typically $<$0.55). As for \textit{Cor}, the trend seems clearer. The \textit{Cor}s of SOTA models in image editing tasks are from 0.3 to 0.5, falling far behind the references. While in image generation, when proprietary models gain \textit{Cor} near 0.8-0.9, almost all open-source models are below 0.5, and open-source UMMs do not show obvious advantages compared with generation-only models. 

We also discover that a wide range of models perform poorly on \textit{Vis} and \textit{Ef}, suggesting that visual incoherence and inconsistency, and unwanted parts in images occur frequently. These observations across sub-tasks and model tiers confirm that our benchmark is challenging for models. 

\circletwo \textbf{ Performance inconsistency between generation and editing tasks.}  A cross-comparison of Table~\ref{tab:main_edit} and Table~\ref{tab:main_gen} reveals a patent performance inconsistency in models, where image editing lags behind image. Empirical evidence shows a significant degradation in both \textit{Cor} and overall average scores across nearly all tasks. For instance, Qwen-Image models achieve \textit{Cor} of 0.46, 0.77, and 0.77 in the generation tasks, but get no more than 0.15 in the editing tasks, while Seedream-4.5 experiences a decline from 0.88 to 0.67 in overall score. This performance gap suggests that while current models, including UMMs, are adept at mapping high-level semantics to visual pixels for generation tasks, they struggle to handle the ``dual-constraint'' of task in editing tasks. Unlike generation, which allows for higher stochastic variance, editing needs fine-grained perception and a complex modifying process, representing higher complexity. 

\circlethree \textbf{ Performance inconsistency between sub-tasks.} Route planning in generation is much more difficult than the other two sub-tasks, as the best \textit{Cor} is only 0.81, compared to 0.90+ of other sub-tasks. While in image editing, all sub-tasks are challenging, as the SOTA scores are much lower. Other models are also sensitive to the categories of sub-tasks. These are likely
due to the mismatch between the models’ mostly reactive
generation behavior and the explicit planning requirements
imposed by different tasks. 

\begin{table}[t]
    \centering
    \setlength\tabcolsep{3pt}
    \caption{Performance comparison between open-ended questions and closed-ended questions. Each score is the average of three sub-tasks. Best correctness is \textbf{\textcolor{red}{highlighted}}.}
    \vspace{-2mm}
    \resizebox{\linewidth}{!}{
    \begin{tabular}{l|ccc|ccc|ccc|ccc}
    \toprule
     \multirow{2}{*}{\textbf{Models}}  & \multicolumn{3}{c|}{Open (Gen)} & \multicolumn{3}{c|}{Closed (Gen)} & \multicolumn{3}{c|}{Open (Edit)} & \multicolumn{3}{c}{Closed (Edit)} \\
    \cmidrule{2-13}
         & \textit{Cor}  & \textit{Vis} & \textit{Ef}  & \textit{Cor}  & \textit{Vis}  & \textit{Ef} & \textit{Cor}  & \textit{Vis}  & \textit{Ef}  & \textit{Cor} & \textit{Vis}  & \textit{Ef}  \\
    \midrule
   \rowcolor{cyan!5} FLUX.1 dev & 0.20 & 0.71 & 0.39 & 0.67 & 0.79 & 0.63 & 0.05 & 0.55 & 0.39 & 0.05 & 0.47 & 0.31 \\
   Ovis-U1 & 0.14 & 0.58 & 0.40 & 0.23 & 0.50 & 0.93 & 0.05 & 0.34 & 0.30 & 0.14 & 0.35 & 0.19 \\
   \rowcolor{cyan!5} OmniGen2 & 0.23 & 0.66 & 0.47 & 0.78 & 0.76 & 0.60 & 0.14 & 0.79 & 0.58 & 0.08 & 0.75 & 0.60 \\
   Bagel & 0.11 & 0.66 & 0.43 & 0.47 & 0.61 & 0.45 & 0.08 & 0.08 & 0.14 & 0.09 & 0.17 & 0.07 \\
   \rowcolor{cyan!5} GPT-Image-1 & \textbf{\textcolor{red}{0.83}} & 0.94 & 0.94 & \textbf{\textcolor{red}{0.96}} & 1.00 & 0.93 & \textcolor{red}{\textbf{0.39}} & 0.55 & 0.75 & \textcolor{red}{\textbf{0.44}} & 0.84 & 0.78 \\
    \bottomrule
    \end{tabular}
    }
    \label{tab:open_close}
    \vspace{-4mm}
\end{table}

\begin{figure}[t]
    \centering
    \includegraphics[width=0.45\linewidth]{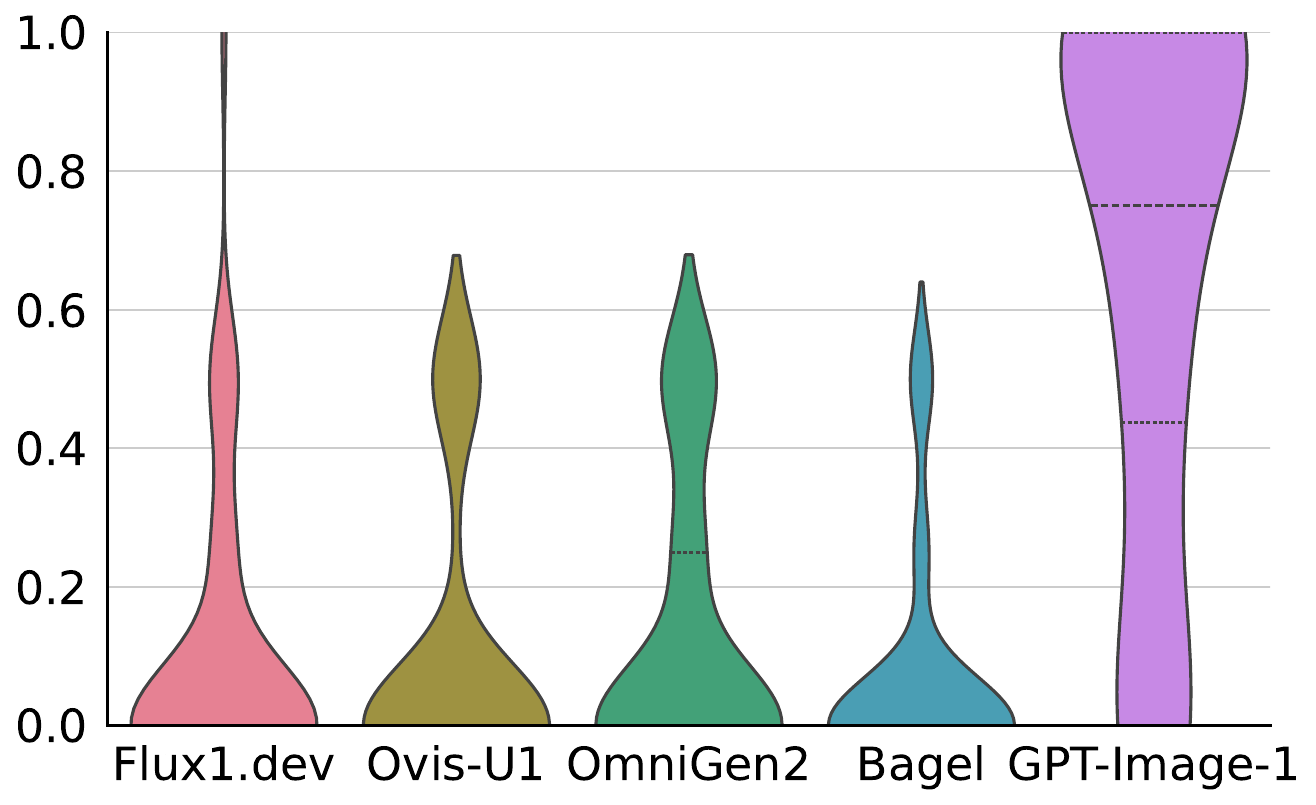}
    \hfill
    \includegraphics[width=0.45\linewidth]{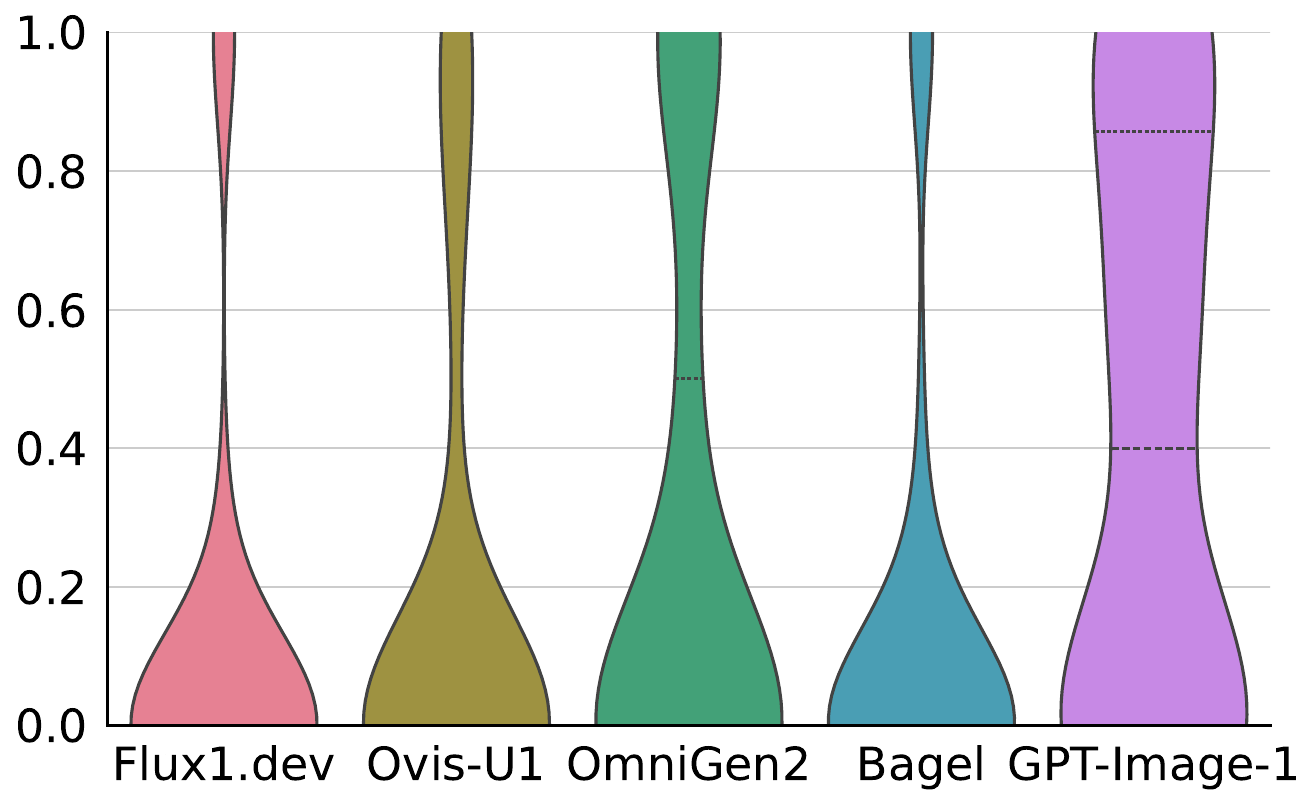}
    \hfill
    \includegraphics[width=0.45\linewidth]{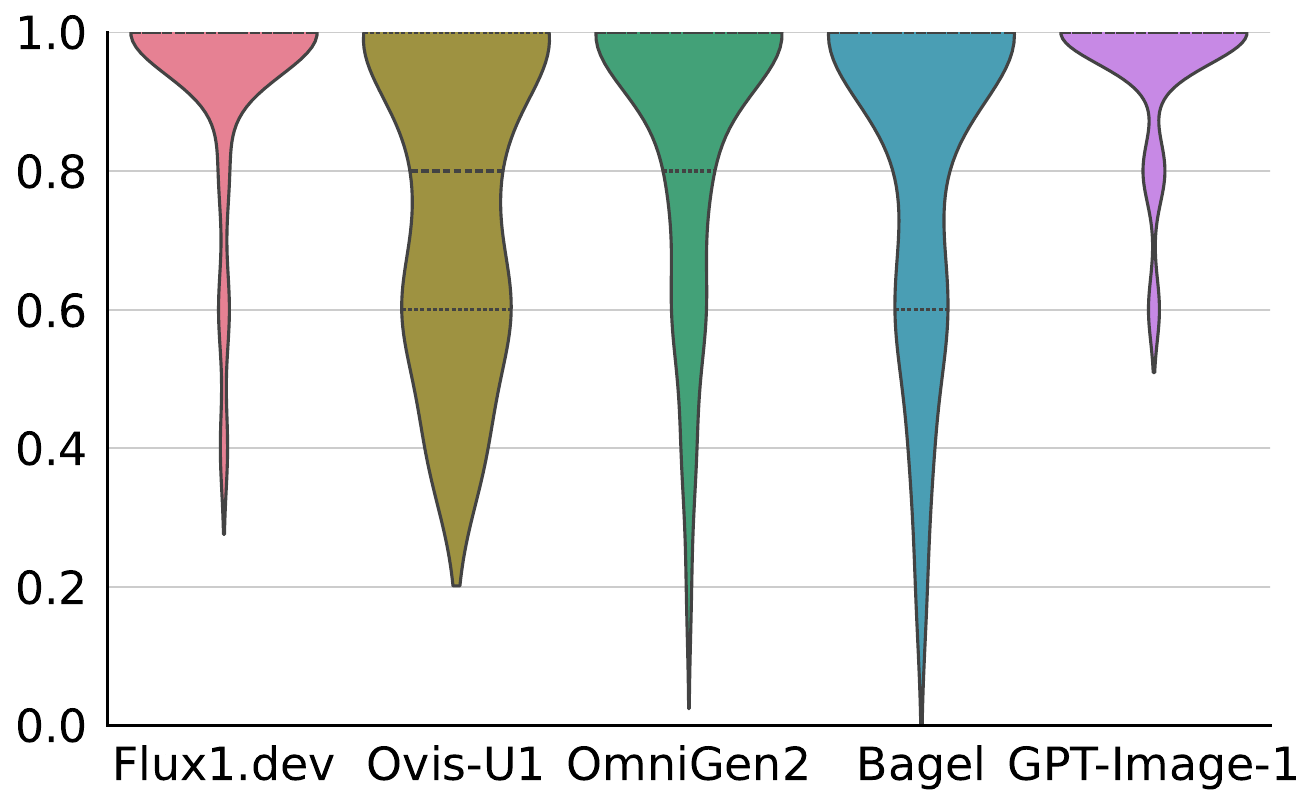}
    \hfill
    \includegraphics[width=0.45\linewidth]{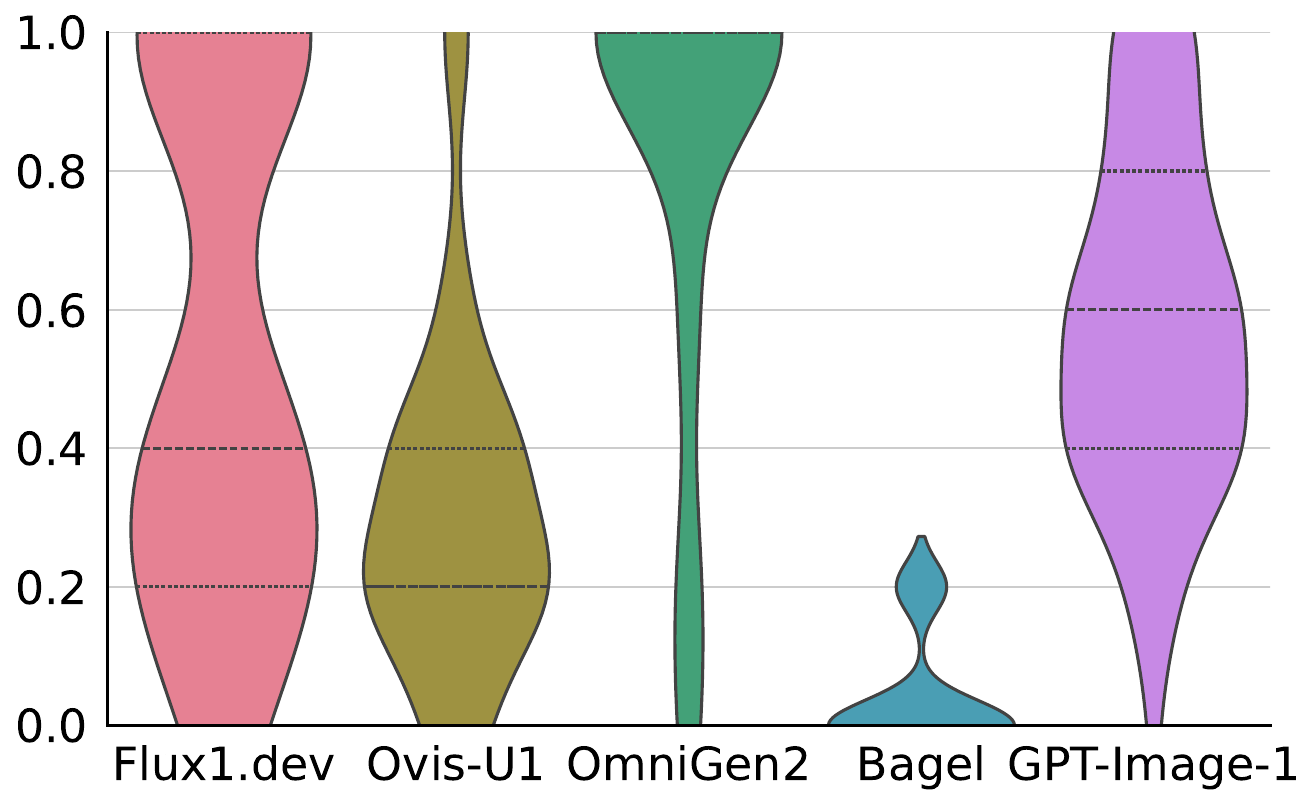}
    
    \caption{Score distribution across different models. We choose \textit{Cor} (top) and \textit{Vis} (bottom) of route planning. }
    \label{fig:score_distribution}
    \vspace{-5mm}
\end{figure}

\begin{figure*}[t]
\centering
\begingroup
\fontsize{6.5}{7.3}\selectfont
\setlength{\tabcolsep}{0pt}
\setlength{\CSUniformSize}{0.097\linewidth}
\begin{tabular}{@{}p{0.097\linewidth}@{\hspace{0.002\linewidth}}p{0.097\linewidth}@{\hspace{0.002\linewidth}}p{0.097\linewidth}@{\hspace{0.002\linewidth}}p{0.097\linewidth}@{\hspace{0.014\linewidth}}p{0.097\linewidth}@{\hspace{0.002\linewidth}}p{0.097\linewidth}@{\hspace{0.002\linewidth}}p{0.097\linewidth}@{\hspace{0.002\linewidth}}p{0.097\linewidth}@{\hspace{0.002\linewidth}}p{0.097\linewidth}@{\hspace{0.002\linewidth}}p{0.097\linewidth}@{}}
\CSUniformCell{FLUX.1-dev~\cite{blackforestlabs2024flux}}{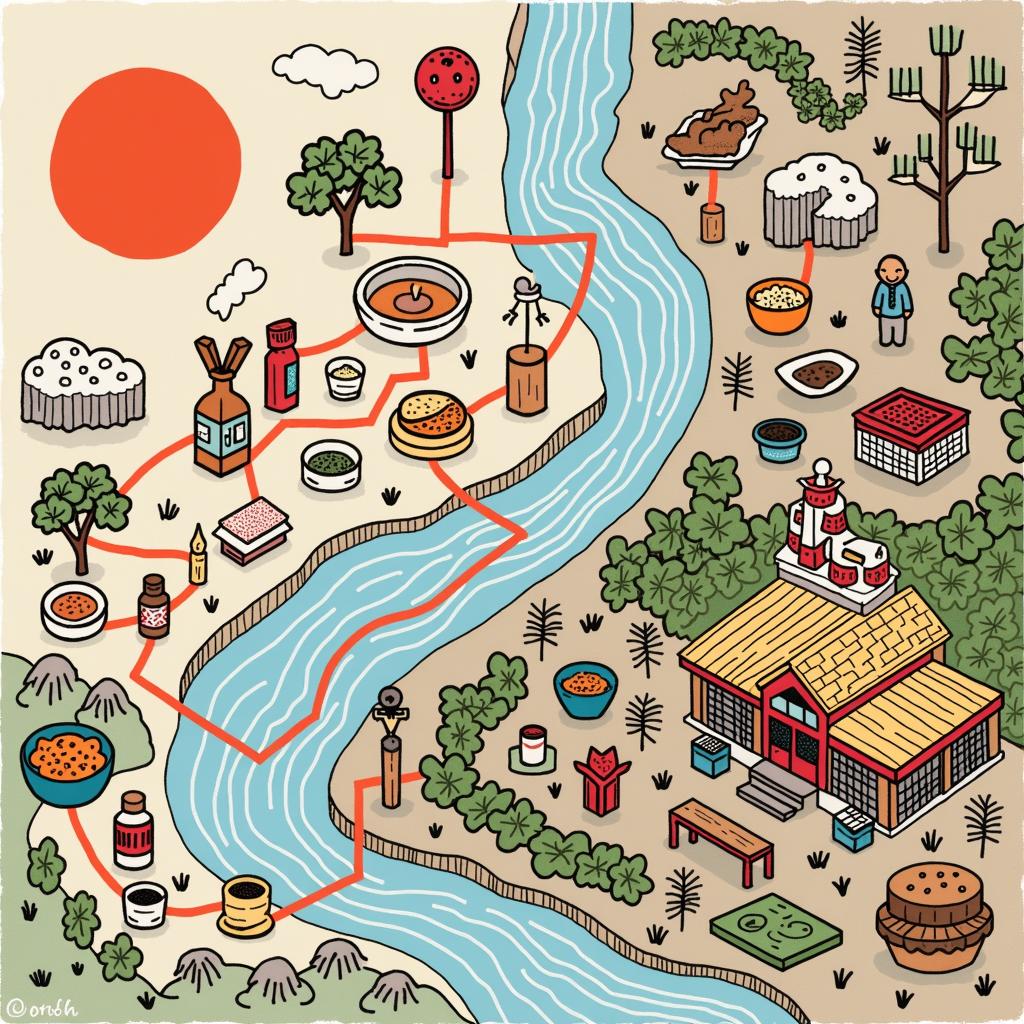} &
\CSUniformCell{UniPic2~\cite{wei2025skywork}}{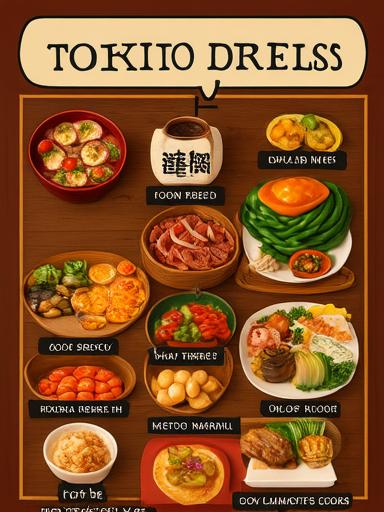} &
\CSUniformCell{\shortstack{Gemini3-Pro-\\Image~\cite{google2025nanopro}}}{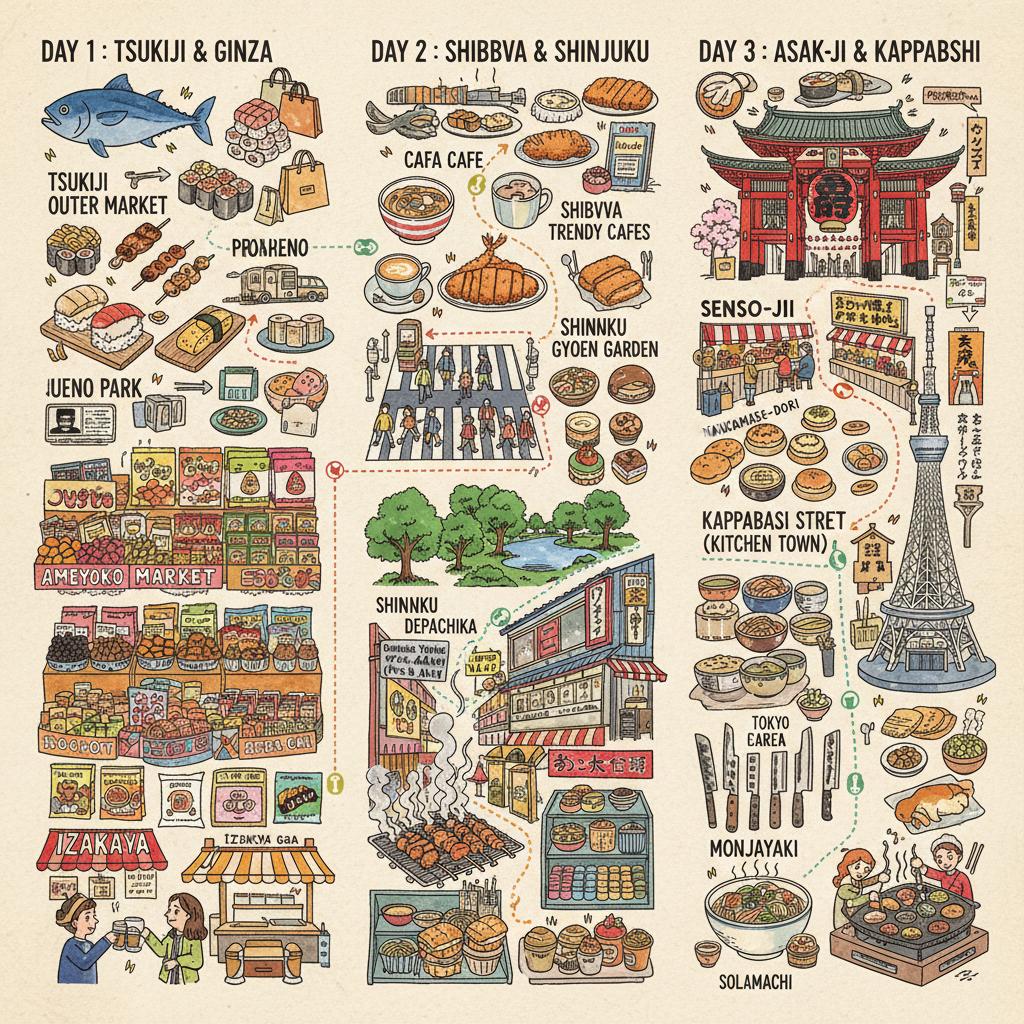} &
\CSUniformCell{GPT-Image-1~\cite{openai2025b_gptimage1}}{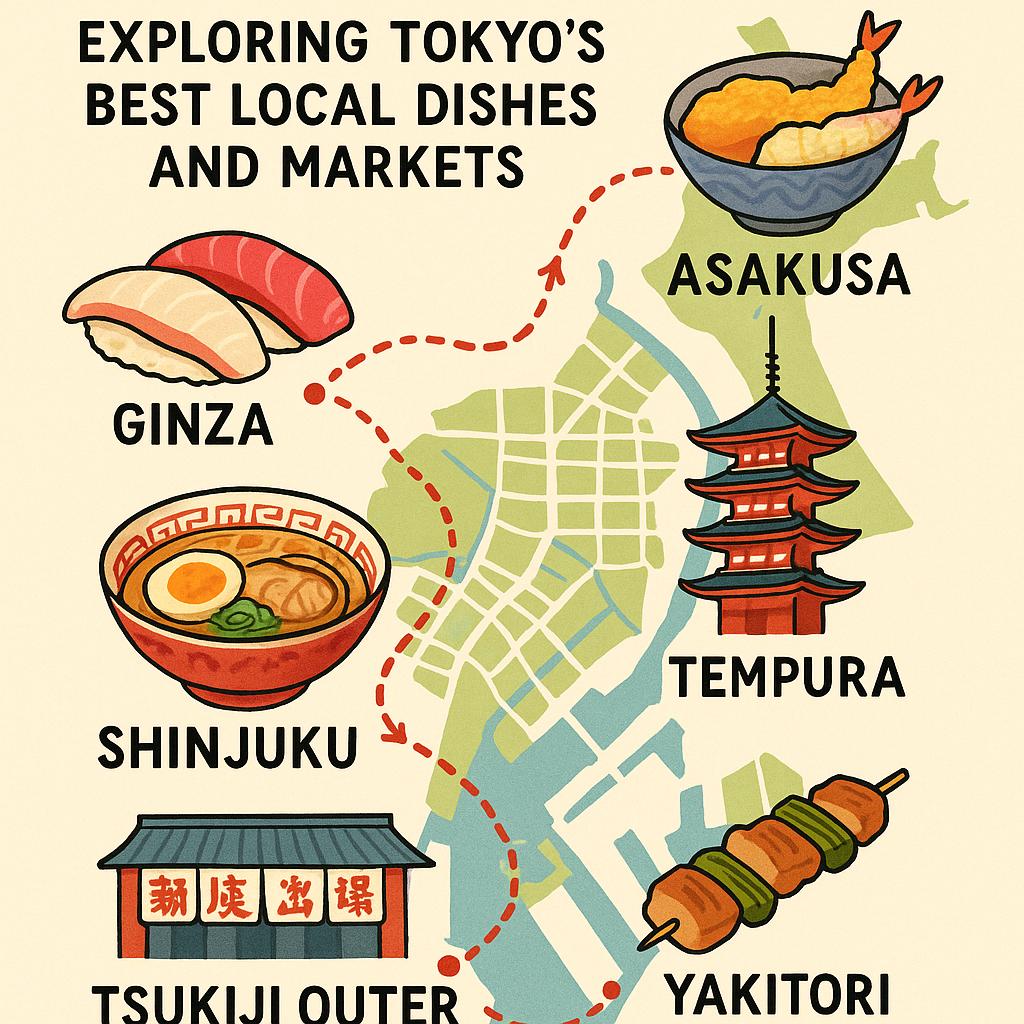} &
\CSUniformCell{Original}{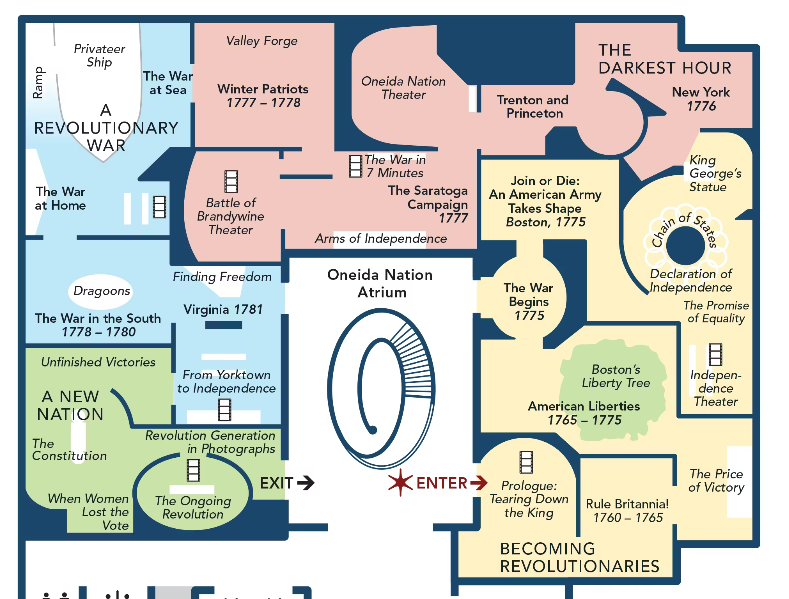} &
\CSUniformCell{Reference}{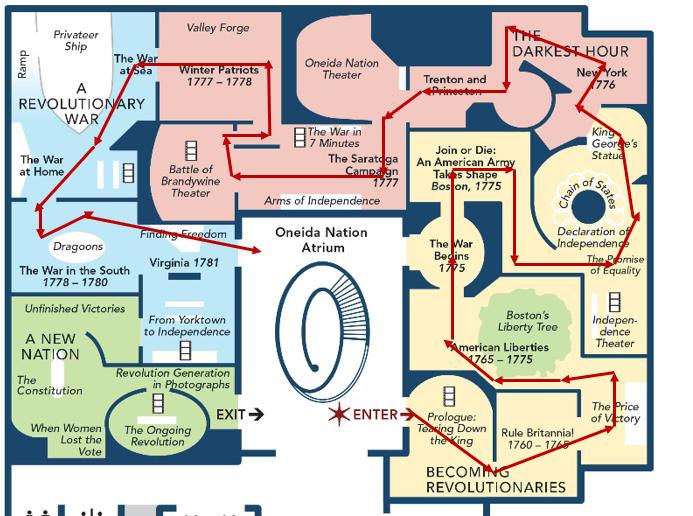} &
\CSUniformCell{Bagel~\cite{deng2025bagel}}{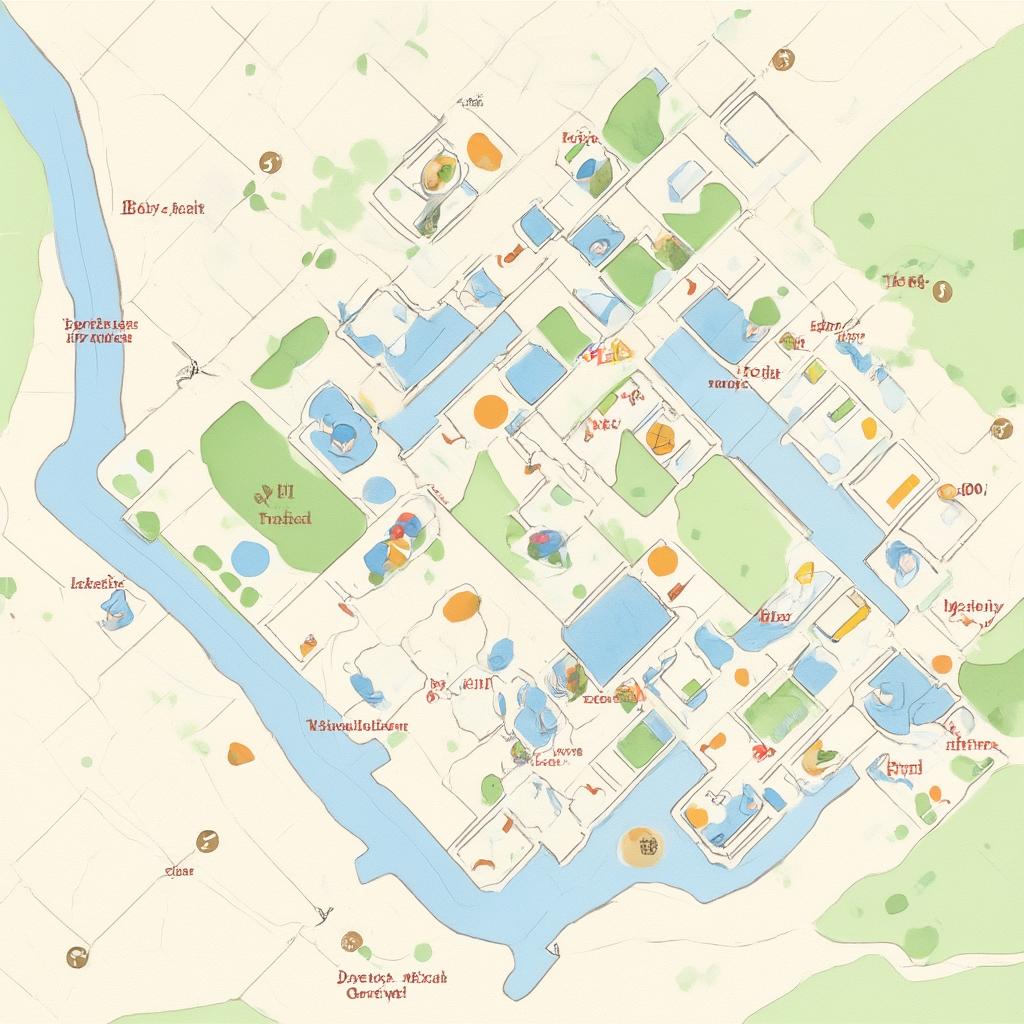} &
\CSUniformCell{Step1X-Edit~\cite{liu2025step1x}}{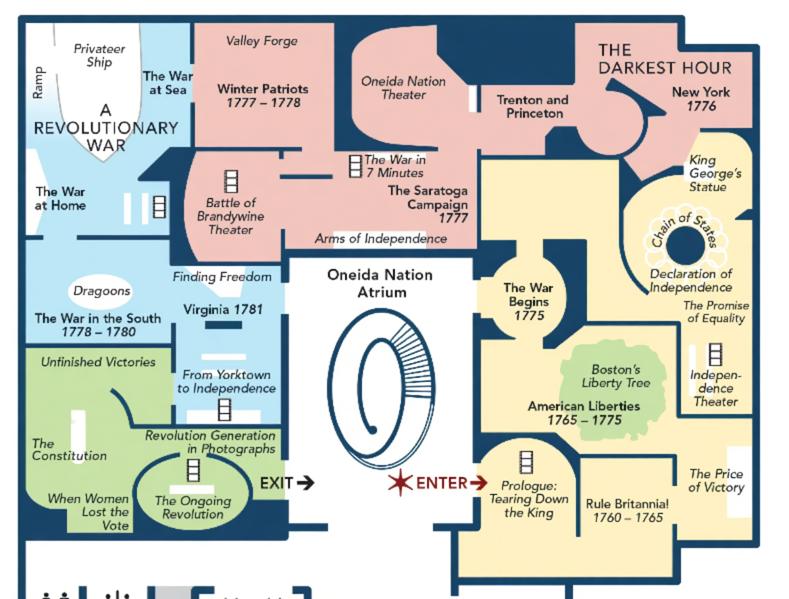} &
\CSUniformCell{\shortstack{Gemini3-Pro-\\Image~\cite{google2025nanopro}}}{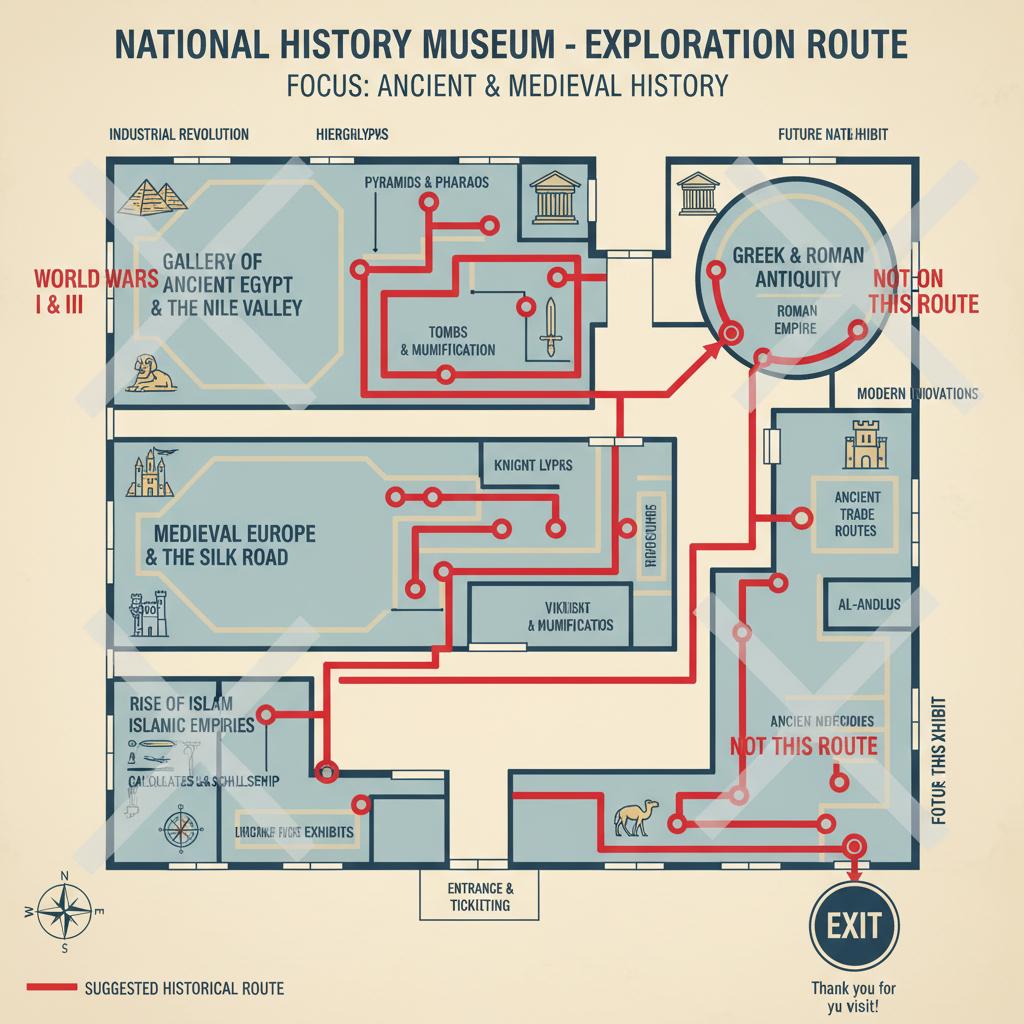} &
\CSUniformCell{GPT-Image-1~\cite{openai2025b_gptimage1}}{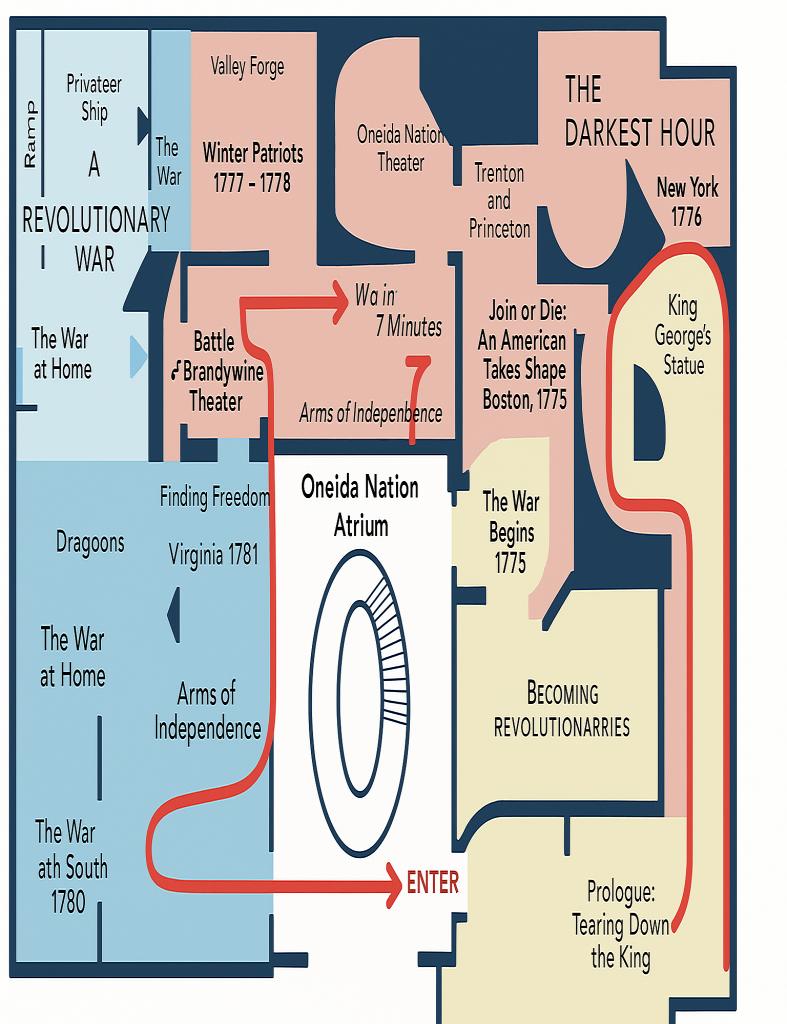} \\
\end{tabular}
\par
\CSPairedPrompts{Can you design a food-focused travel route for exploring Tokyo's best local dishes and markets and draw it?}{Create an exploratory route that covers at least six exhibition halls on this map.}
\begin{tabular}{@{}p{0.097\linewidth}@{\hspace{0.002\linewidth}}p{0.097\linewidth}@{\hspace{0.002\linewidth}}p{0.097\linewidth}@{\hspace{0.002\linewidth}}p{0.097\linewidth}@{\hspace{0.014\linewidth}}p{0.097\linewidth}@{\hspace{0.002\linewidth}}p{0.097\linewidth}@{\hspace{0.002\linewidth}}p{0.097\linewidth}@{\hspace{0.002\linewidth}}p{0.097\linewidth}@{\hspace{0.002\linewidth}}p{0.097\linewidth}@{\hspace{0.002\linewidth}}p{0.097\linewidth}@{}}
\CSUniformCell{Bagel}{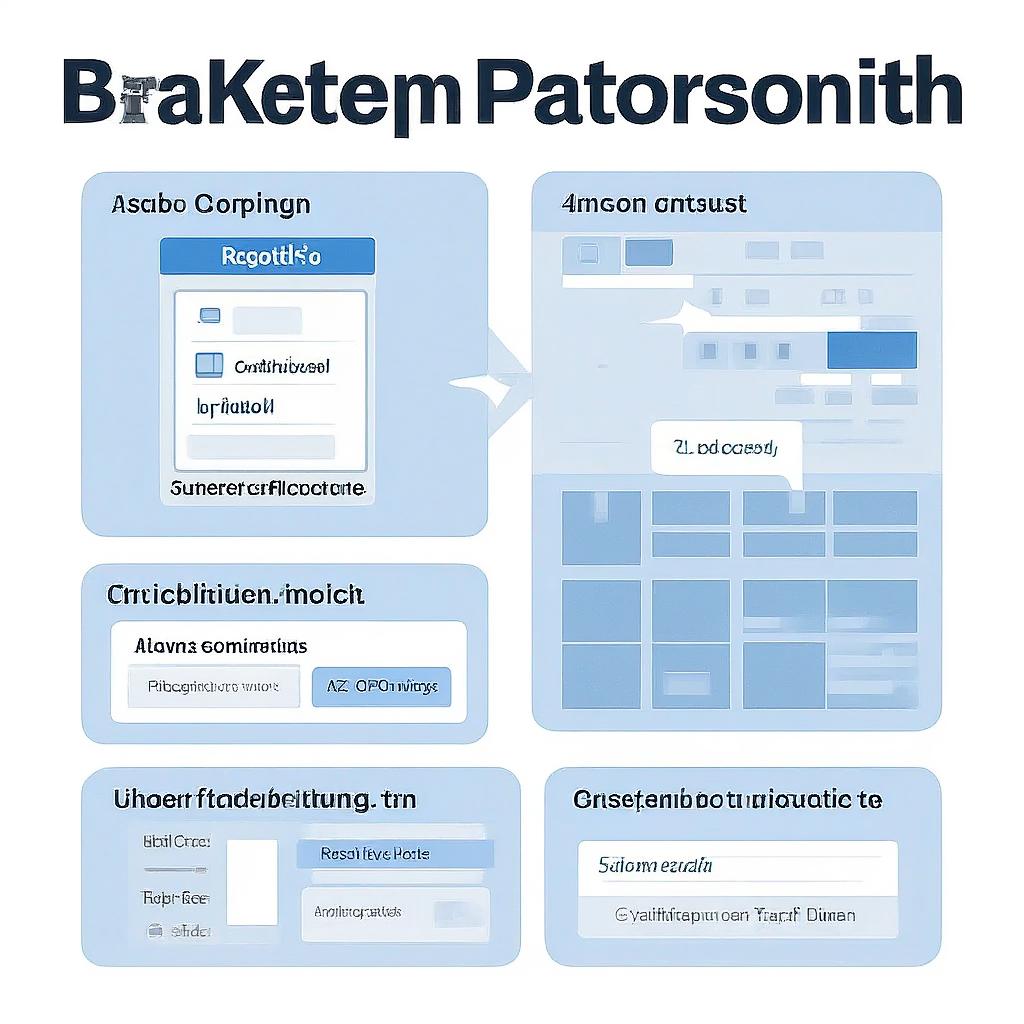} &
\CSUniformCell{Janus-4o~\cite{chen2025sharegpt}}{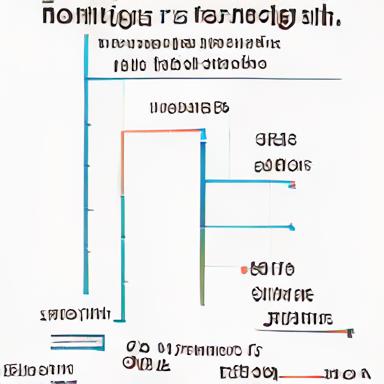} &
\CSUniformCell{\shortstack{Gemini3-Pro-\\Image~\cite{google2025nanopro}}}{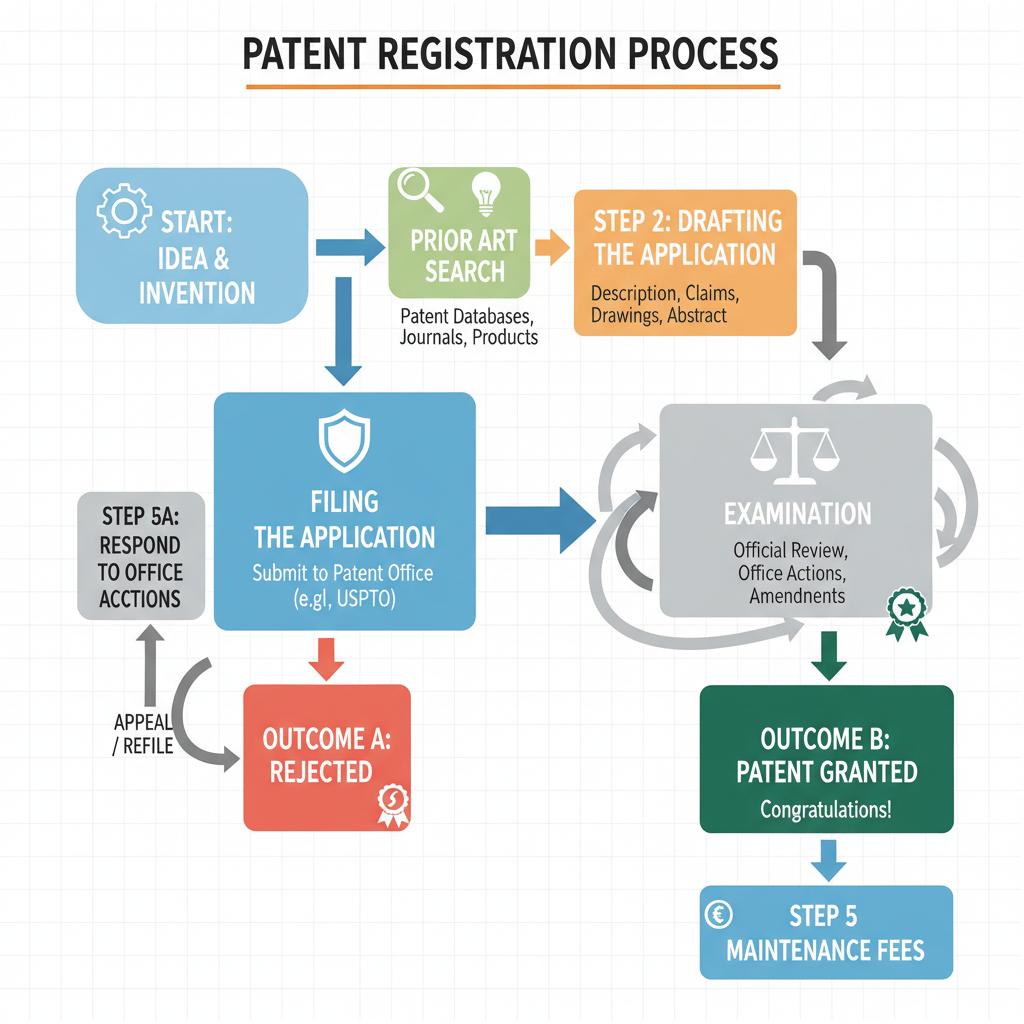} &
\CSUniformCell{GPT-Image-1~\cite{openai2025b_gptimage1}}{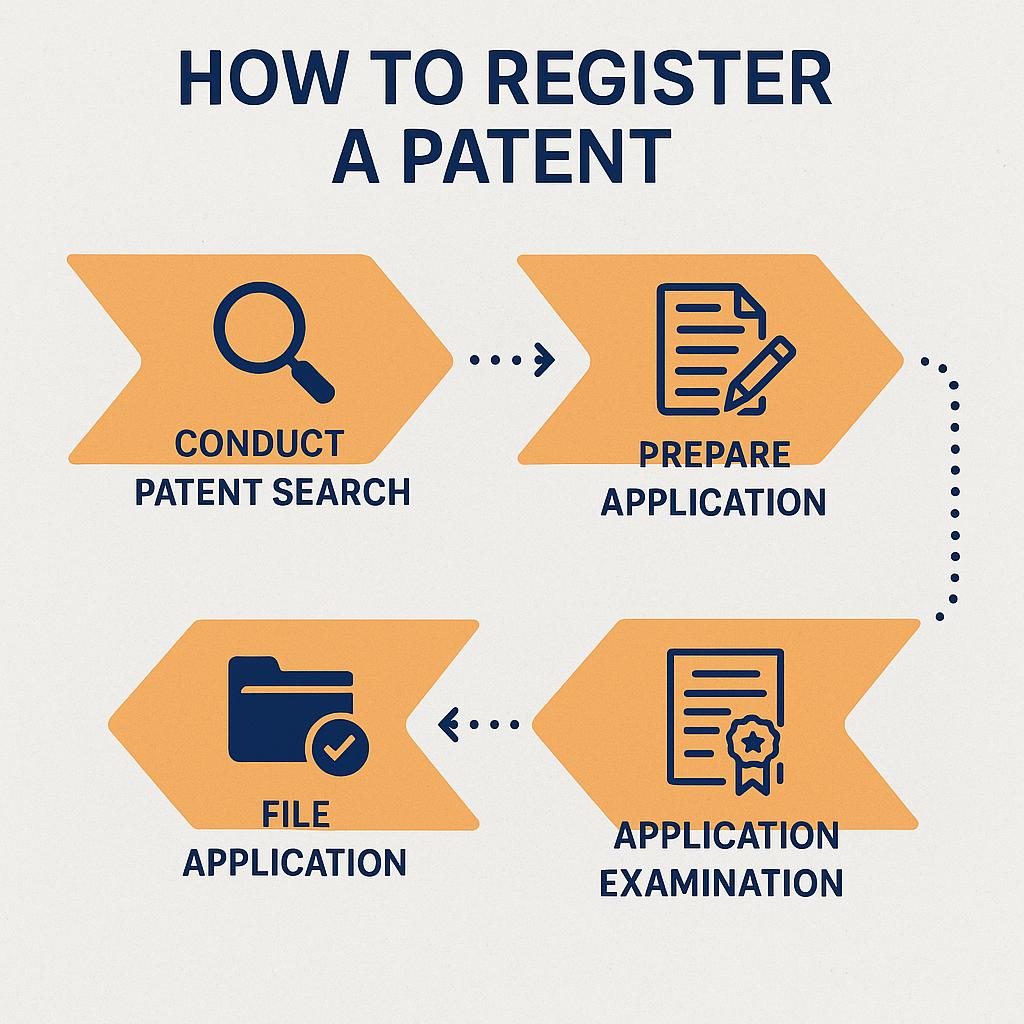} &
\CSUniformCell{Original}{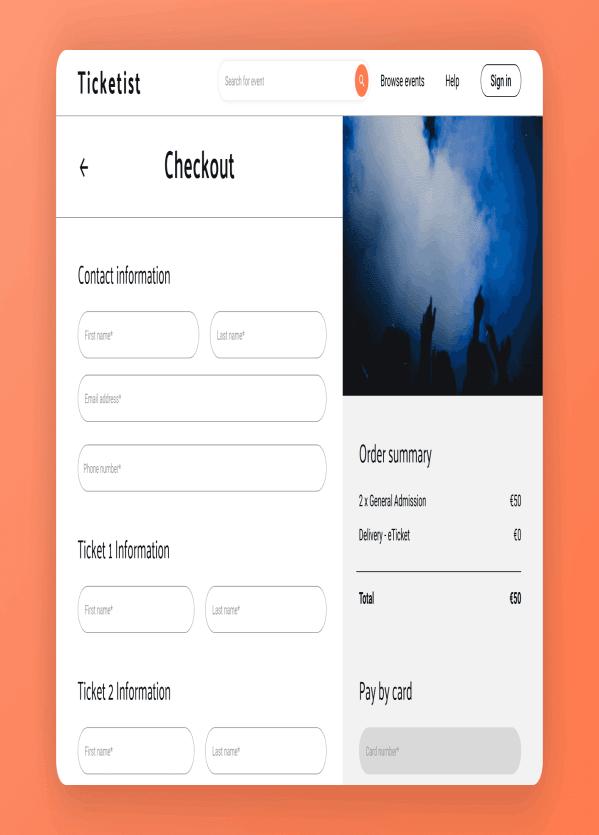} &
\CSUniformCell{Reference}{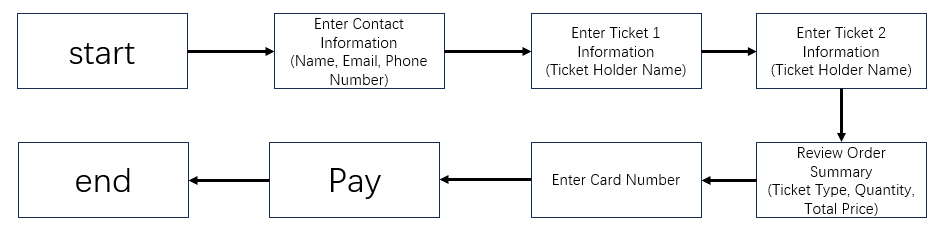} &
\CSUniformCell{Bagel~\cite{deng2025bagel}}{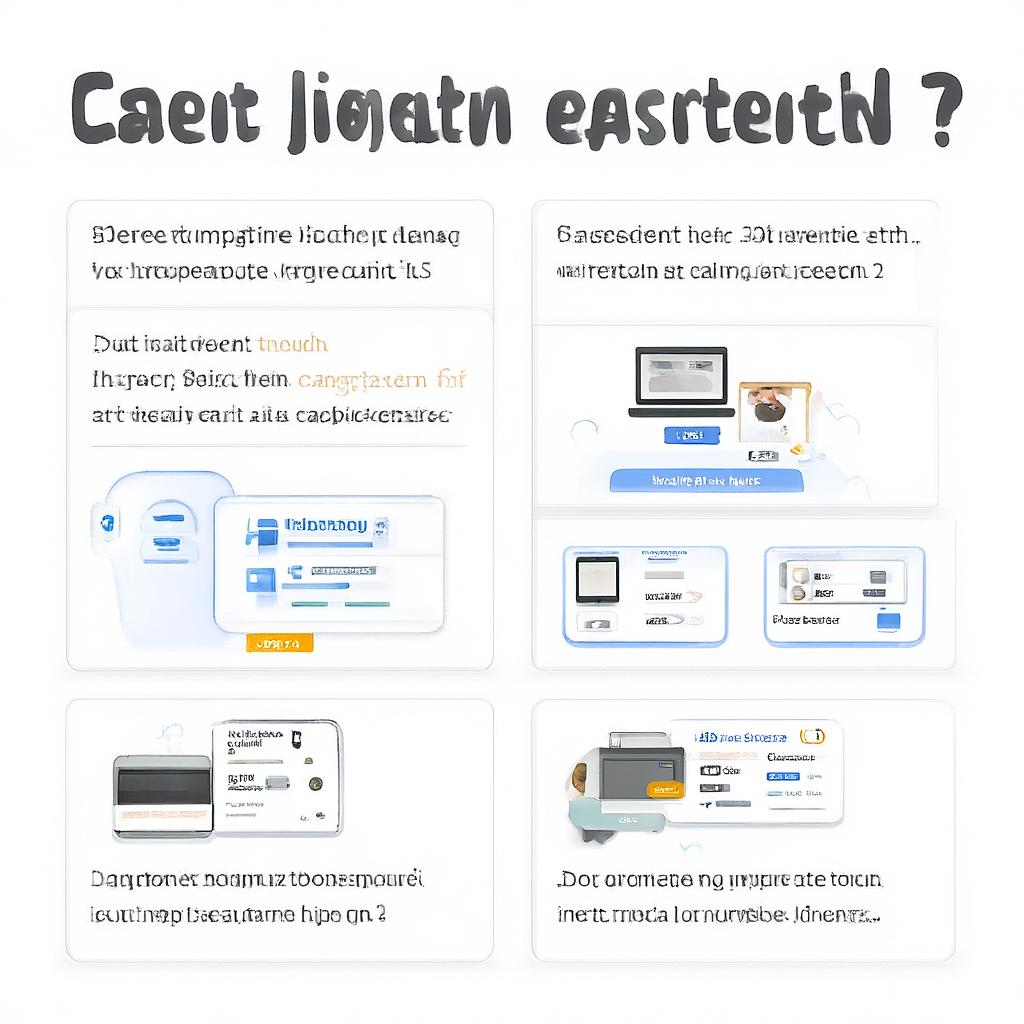} &
\CSUniformCell{Janus-4o~\cite{chen2025sharegpt}}{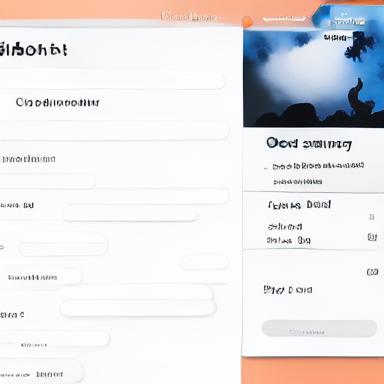} &
\CSUniformCell{\shortstack{Gemini3-Pro-\\Image~\cite{google2025nanopro}}}{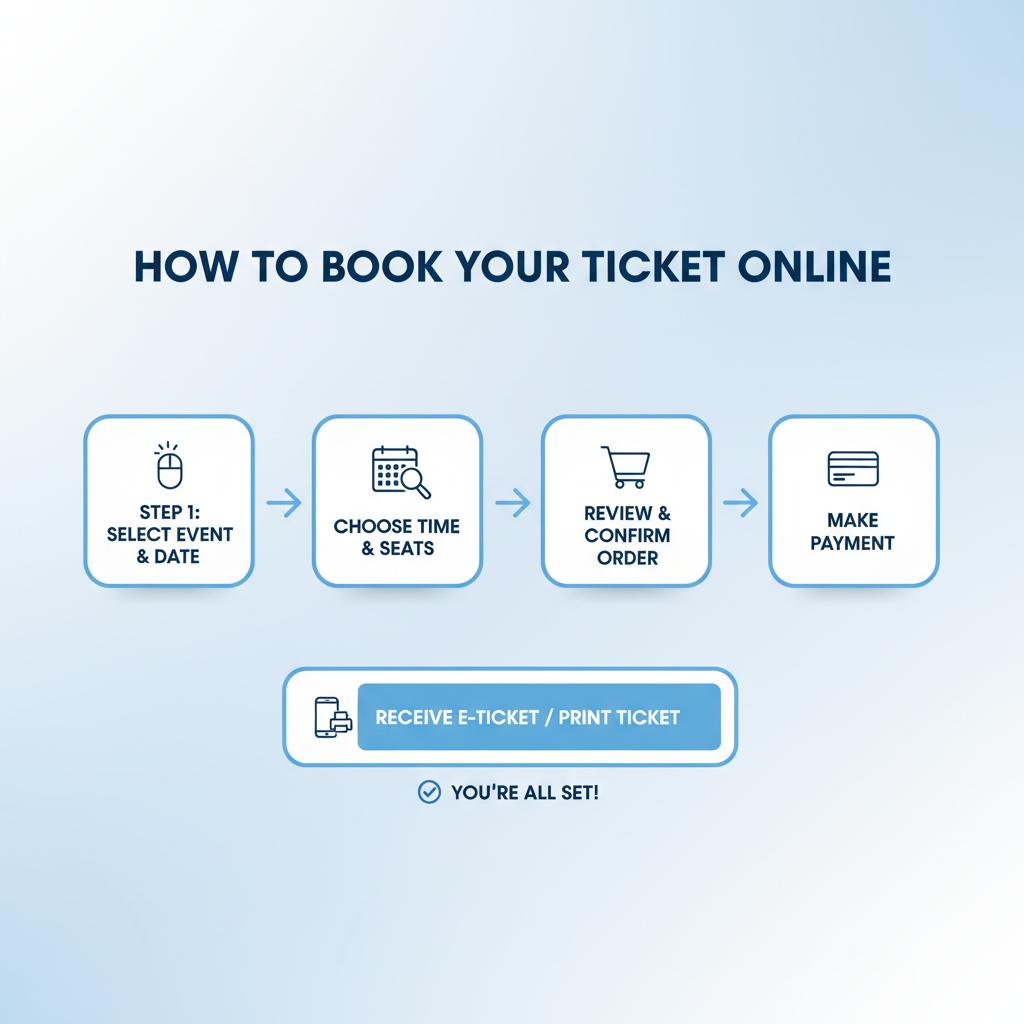} &
\CSUniformCell{GPT-Image-1~\cite{chen2025sharegpt}}{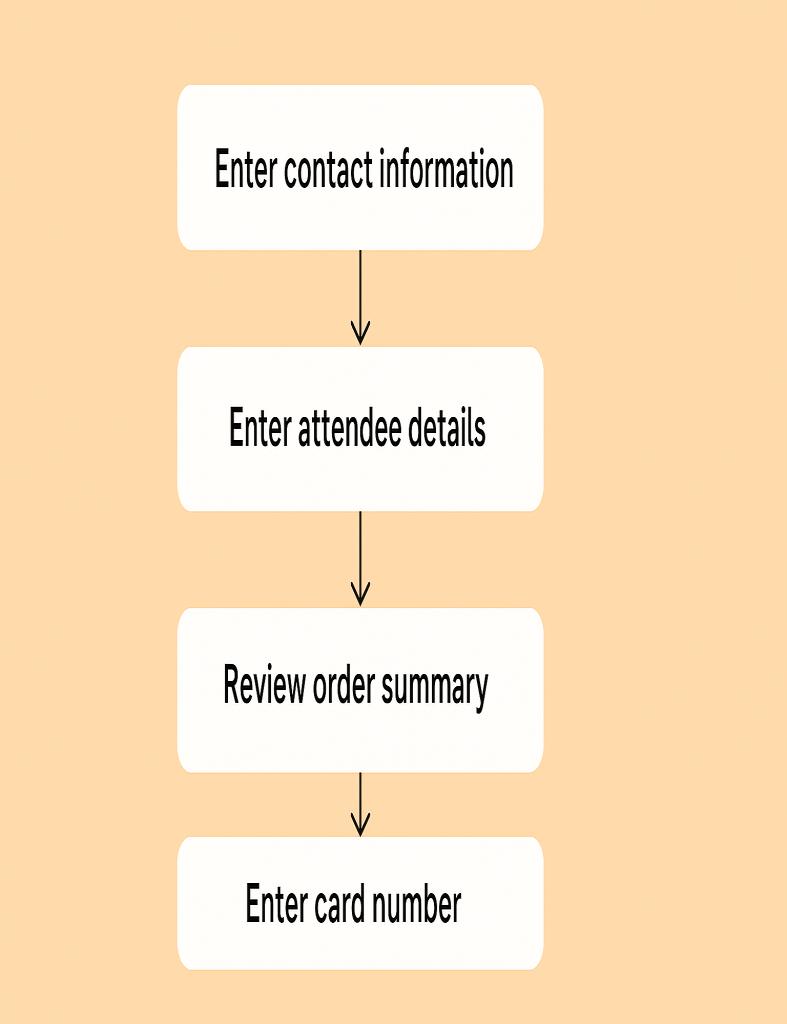} \\
\end{tabular}
\par
\CSPairedPrompts{Design a process-oriented visual chart to explain how to registering a patent.}{Design a process-oriented visual chart to explain how to book a ticket on this page.}
\begin{tabular}{@{}p{0.097\linewidth}@{\hspace{0.002\linewidth}}p{0.097\linewidth}@{\hspace{0.002\linewidth}}p{0.097\linewidth}@{\hspace{0.002\linewidth}}p{0.097\linewidth}@{\hspace{0.014\linewidth}}p{0.097\linewidth}@{\hspace{0.002\linewidth}}p{0.097\linewidth}@{\hspace{0.002\linewidth}}p{0.097\linewidth}@{\hspace{0.002\linewidth}}p{0.097\linewidth}@{\hspace{0.002\linewidth}}p{0.097\linewidth}@{\hspace{0.002\linewidth}}p{0.097\linewidth}@{}}
\CSUniformCell{Bagel~\cite{deng2025bagel}}{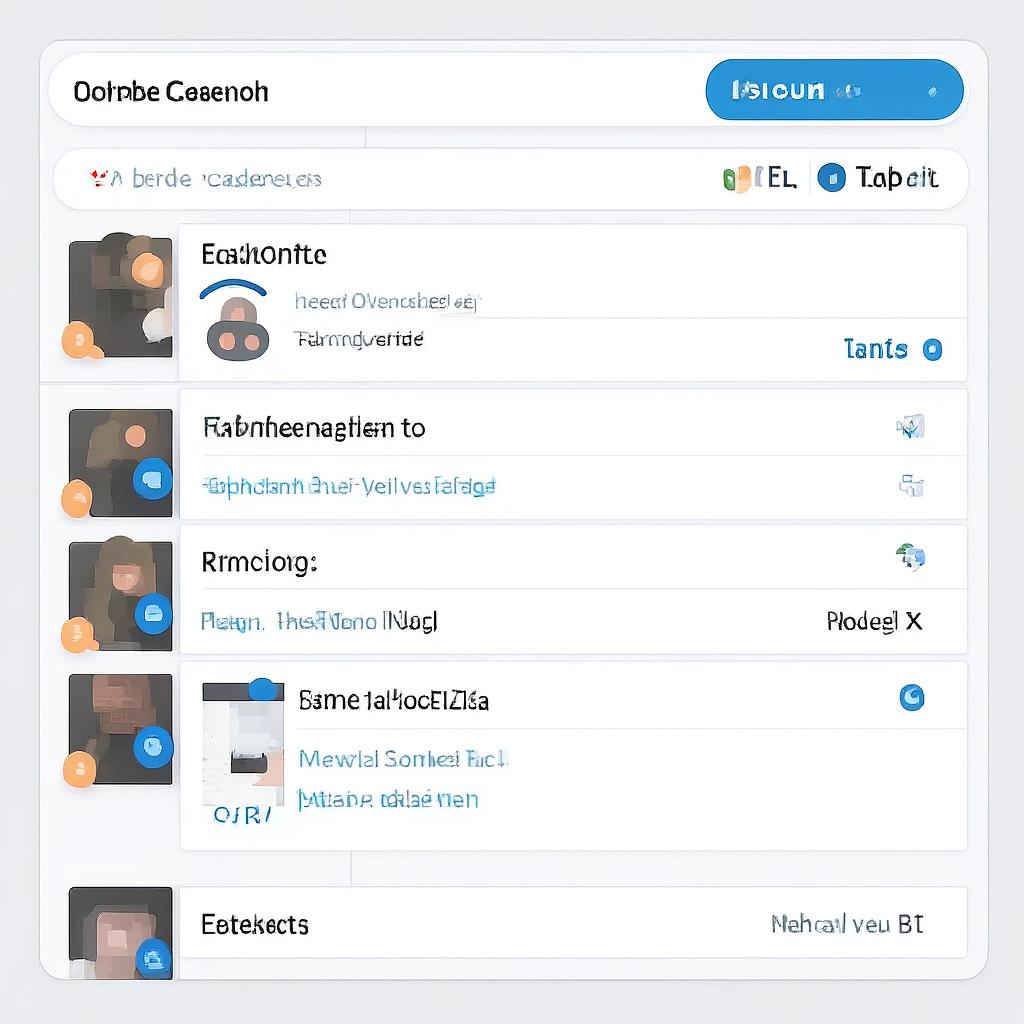} &
\CSUniformCell{Janus-4o~\cite{chen2025sharegpt}}{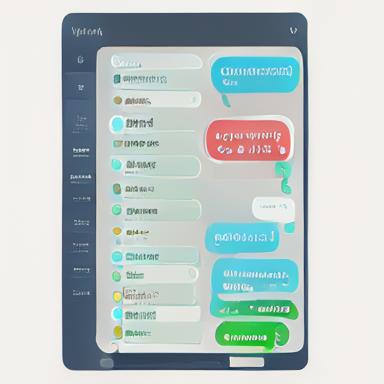} &
\CSUniformCell{\shortstack{Gemini3-Pro-\\Image~\cite{google2025nanopro}}}{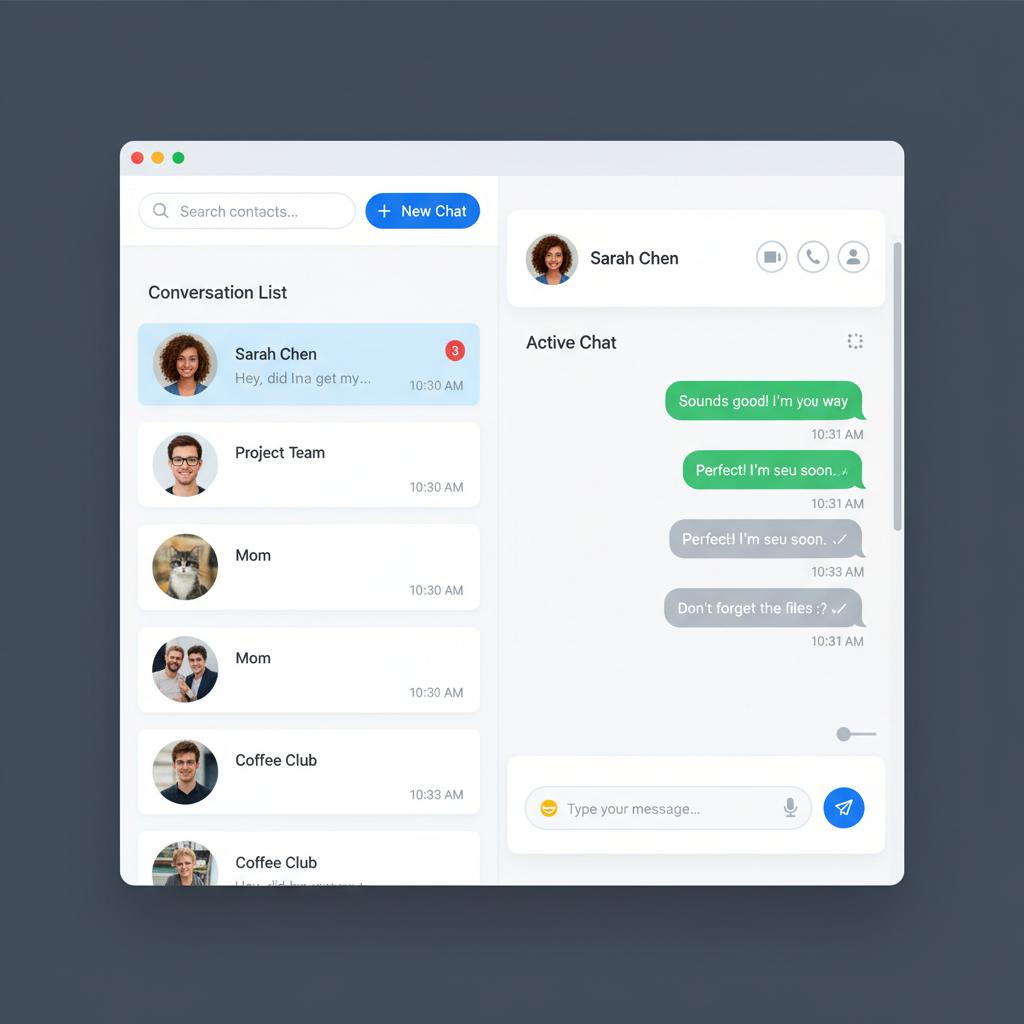} &
\CSUniformCell{GPT-Image-1~\cite{chen2025sharegpt}}{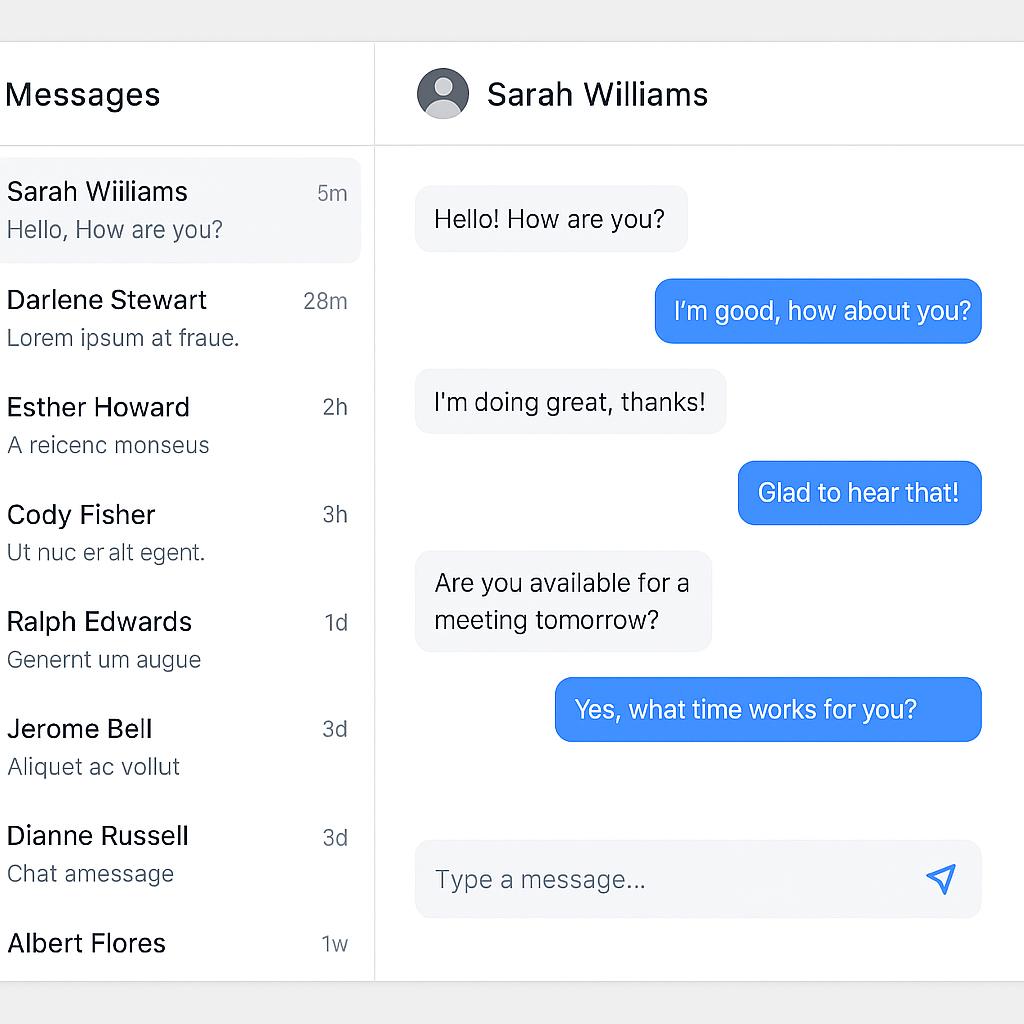} &
\CSUniformCell{Original}{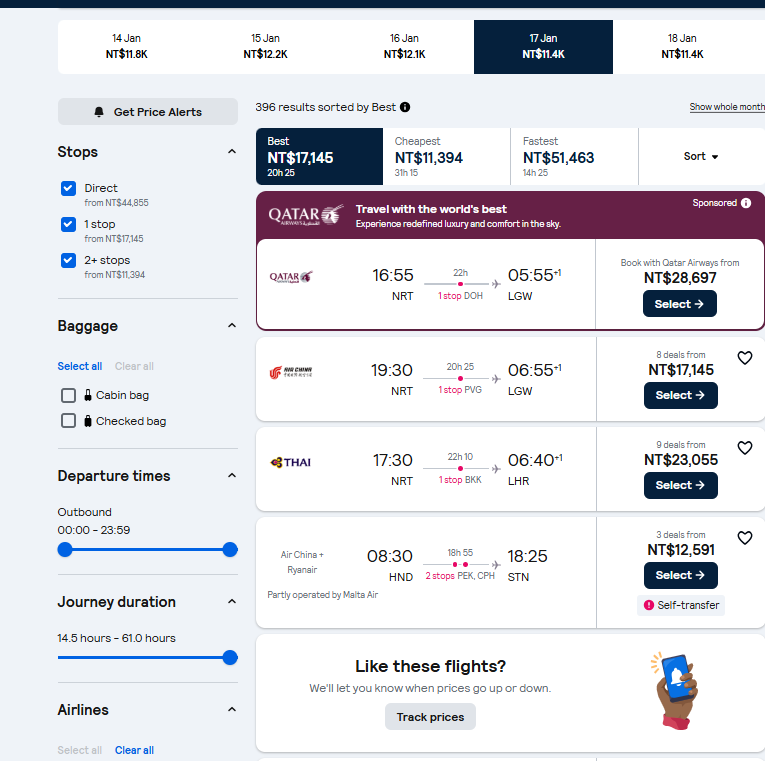} &
\CSUniformCell{Reference}{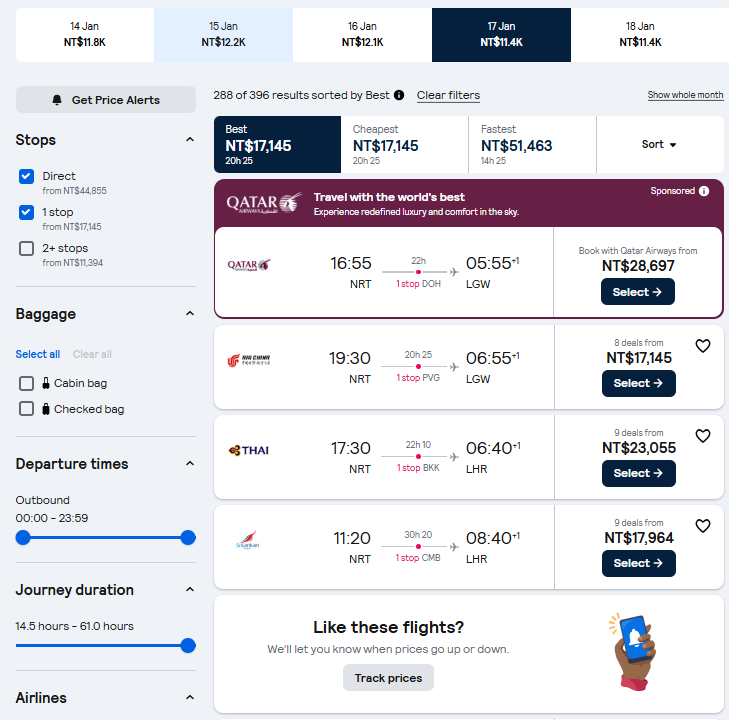} &
\CSUniformCell{Bagel~\cite{deng2025bagel}}{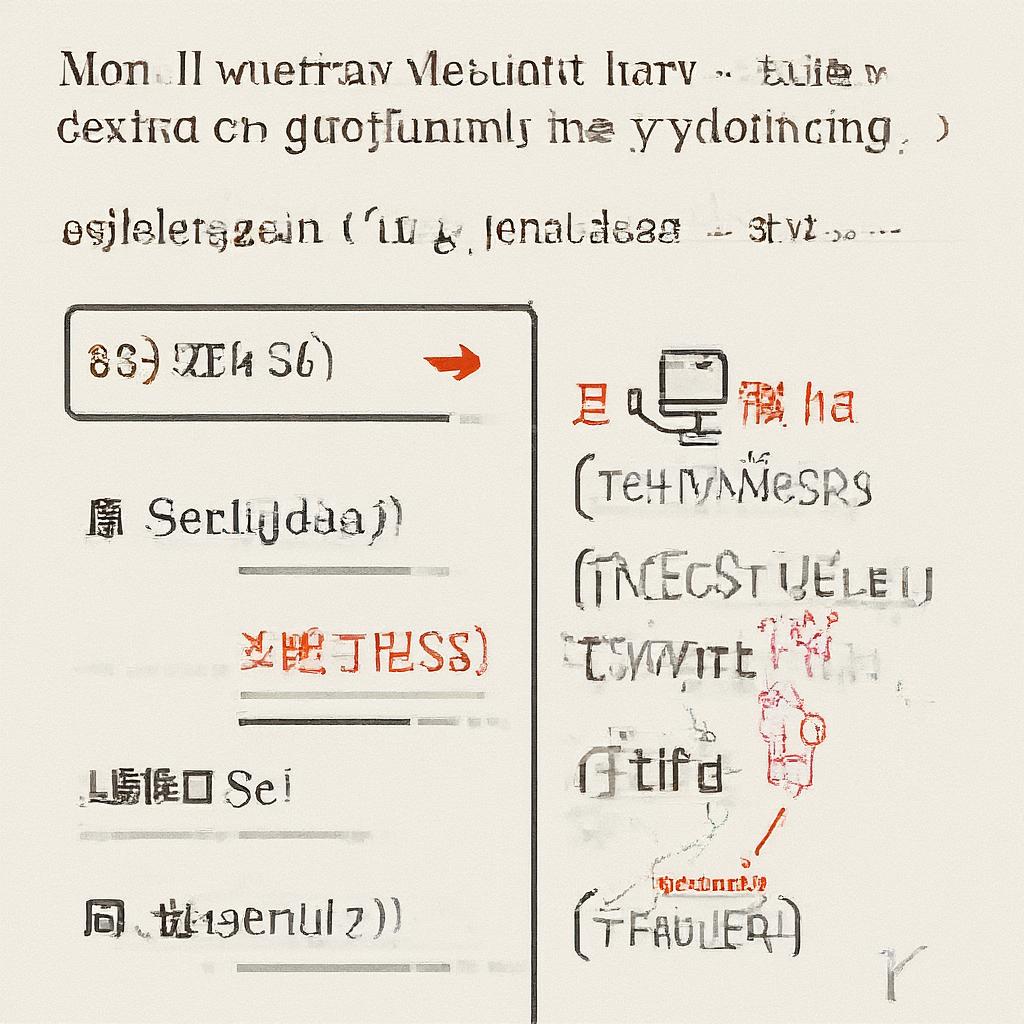} &
\CSUniformCell{Janus-4o~\cite{chen2025sharegpt}}{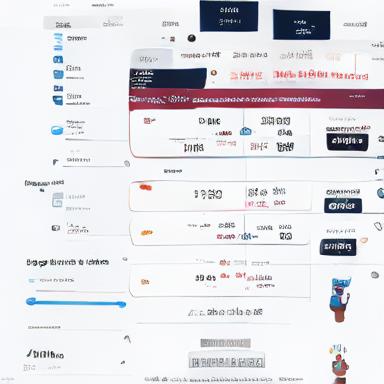} &
\CSUniformCell{\shortstack{Gemini3-Pro-\\Image~\cite{google2025nanopro}}}{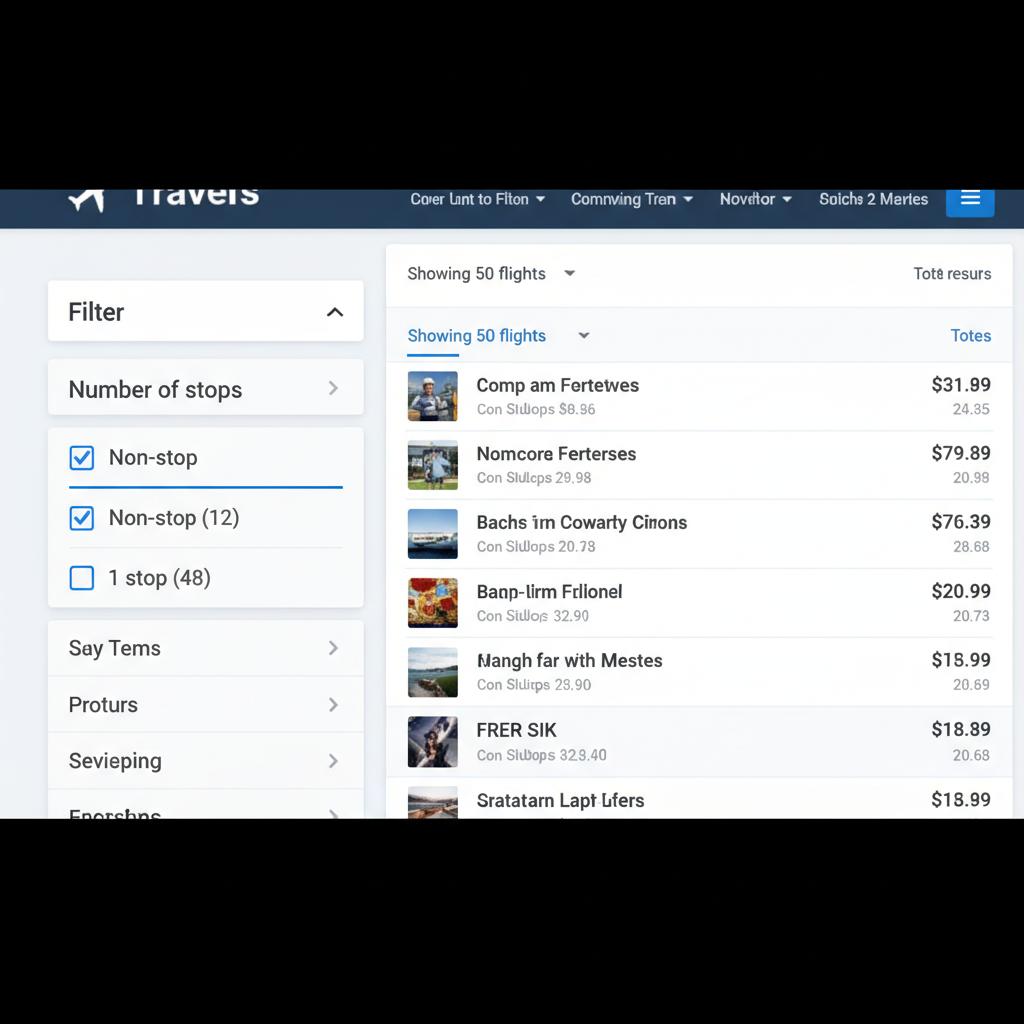} &
\CSUniformCell{GPT-Image-1~\cite{chen2025sharegpt}}{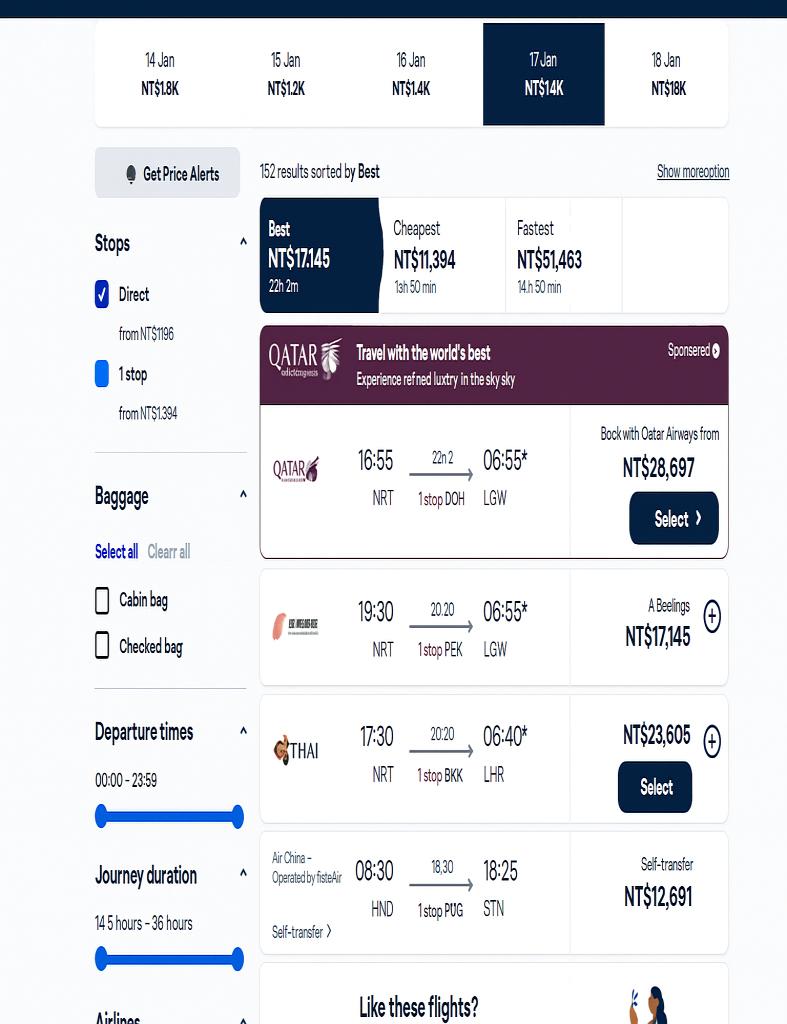} \\
\end{tabular}
\par
\CSPairedPrompts{Design a web UI mockup for a blog article page with the title at the top, content in the center, and a sidebar on the right.}{Draw what will happen to this page when the user cancel the choice of 2+ stops.}
\endgroup
\vspace{-3mm}
\caption{Qualitative examples of Route Planning, Workflow Diagramming, and Web\&UI Displaying (top to bottom). Each row shows Generation on the left and Editing on the right, separated by a small gap. Editing columns show Original, Reference, and outputs.}
\label{fig:case}
\vspace{-3mm}
\end{figure*}

\circlefour \textbf{ Task-specific sensitivity of thinking-based generation.} The integration of ``Thinking" mechanisms, including GRPO~\cite{guo2025deepseek}, within UMMs exhibits non-uniform effectiveness across different task categories in PlanViz. As evidenced in Table~\ref{tab:main_edit}, the Thinking variant of Step1X-Edit-v1p2 shows a marginal improvement in overall average scores (+0.04). For correctness scores, the performance improvement of Step1X-Edit-v1p2 only increases by 0.02-0.04. Models like Bagel and UniPic2-Metaquery see a similar situation, and even in some cases, the thinking mode leads to worse performance, like web\&UI displaying for Bagel and route planning for UniPic2-Metaquery. However, results in Table~\ref{tab:main_gen} reveal a different trend. In image generation tasks, thinking mode improves the \textit{Cor} of bagel on all sub-tasks by up to 0.35, and UniPic2-Metaquery by up to 0.13, much more than image editing tasks. Consequently, thinking-based generation shows task-dependent performances. 

\circlefive \textbf{ Decoupling of \textit{Cor} and other scores.}Tables~\ref{tab:main_edit} and~\ref{tab:main_gen} show that many models maintain high Vis scores ($>$0.70) despite near-zero \textit{Cor}, indicating they frequently generate "visually plausible but semantically mismatch" content. Conversely, several cases that achieve much higher \textit{Cor} often suffer a regression in \textit{Vis} and \textit{Ef}, like Janus-4o in editing tasks of route planning. This trade-off suggests that while these models attempt to adhere to logical constraints, they lack the representational robustness or domain-specific knowledge to maintain visual fluency under strict instruction, resulting in ``correct but degraded'' outputs that hit isolated scoring points but fail in overall synthesis. Consequently, bridging this gap remains the primary challenge for modern UMMs.

\subsubsection{Open-ended v.s. closed-ended}

\label{sec:open_close}

One image-only model, three open-source UMMs, and one closed-source UMM are chosen here to explore the effects on generation and editing. Results are in Table~\ref{tab:open_close}. It's patently obvious that GPT-Image-1 performs better than all open-source models, supporting the findings in Section~\ref{sec:findings}. More importantly, it seems that the \textit{Cor}s of almost all closed-ended cases are better than those of open-ended cases. This may suggest that proper constraints and details can help UMM plan and generate better in this domain. 

\subsubsection{Can code rendering do well?}
\label{sec:code}
\begin{table}[t]
    \centering
    \caption{Performance of code rendering. All evaluation follows the main table. We only count the successfully rendered samples here.}
    \vspace{-2mm}
    \resizebox{\linewidth}{!}{
    \begin{tabular}{c|ccc|ccc}
    \toprule
     \multirow{2}{*}{\textbf{Sub-tasks}}    & \multicolumn{3}{c|}{Gen} & \multicolumn{3}{c}{Edit} \\
         \cmidrule{2-7}
         & \textit{Cor} & \textit{Vis} & \textit{Ef} & \textit{Cor} & \textit{Vis} & \textit{Ef} \\
         \midrule
      \rowcolor{red!5} Route Planning  & 0.99 & 0.99 & 0.94 & 0.75 & 0.83 & 0.93 \\
       Workflow Diagramming  & 1.00 & 0.99 & 0.98 & 0.55 & 0.27 & 0.94 \\
      \rowcolor{red!5} Web \& UI Displaying  & 0.99 & 0.99 & 0.84 & 0.77 & 0.68 & 0.92 \\
       \bottomrule
    \end{tabular}
    }
    \label{tab:code}
    \vspace{-5mm}
\end{table}
\begin{figure}[ht]
    \centering
    \includegraphics[width=\linewidth]{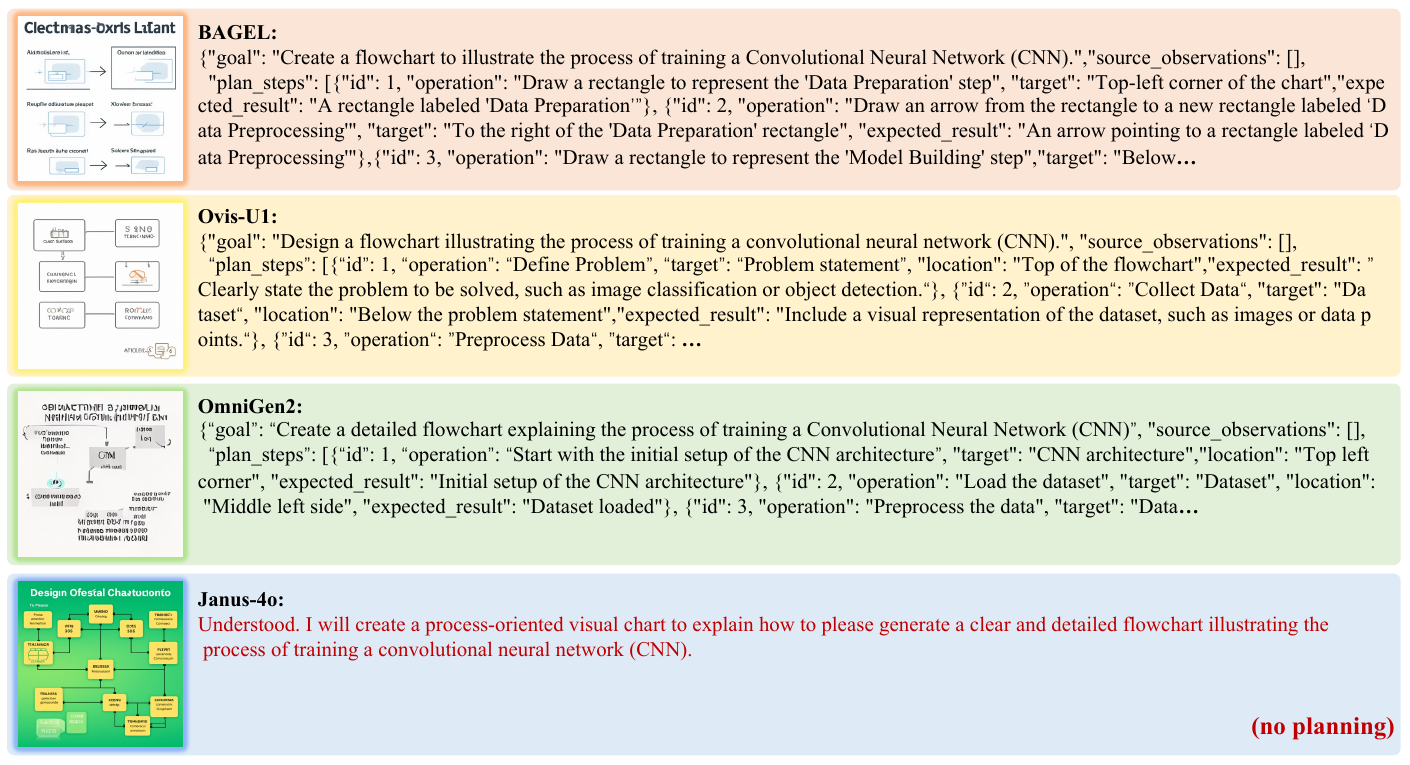}
    \vspace{-3mm}
    \caption{Gap between knowing how to plan and real visualization of UMMs. The query is: \textit{Design a process-oriented visual chart to explain how to please generate a clear and detailed flowchart illustrating the process of training a convolutional neural network (CNN).} We reports the textual planning outputs using understanding mode, and the related generated images in generation mode.}
    \label{fig:gap_plan_gen}
    \vspace{-5mm}
\end{figure}
Since computer-use tasks are evaluated, we naturally explore whether pure code+rendering can do well. We leverage GPT-5.6-sol~\cite{openai2025gpt5} with Codex. Details and cases are in Appendix~\textcolor{cvprblue}{M}. Table~\ref{tab:code} reports the results. 
Code rendering performs strongly on generation tasks, where explicit structural specifications can be directly translated into visual layouts. Its advantage becomes smaller in editing, which additionally requires preserving and reasoning over implicit constraints (like ``the rooms should be suitable and prices should be higher for more travelers'') from the source image. This may need model reasoning, pointing out a promising way. 

\subsubsection{Influence of prompt styles.}
10 questions are randomly sampled and we use GPT-5.1~\cite{openai2025gpt5} to create 10 language styles of each. Then they are given several UMMs for results. Detailed scores can be found in Appendix~\textcolor{cvprblue}{N}. According to that, we discover that (i) open-source UMMs are influenced more than GPT-Image-1, as the standard deviations compared to mean scores are higher (ii) the performance of open-source UMMs varies; however, they often obtain similar scores on the same question. (like $Cor$ of question 5, $Vis$ of question 3, $Ef$ of question 7, etc.). This suggests that open-source UMMs still cannot remain robust to all kinds of input styles under our task.

\subsubsection{Distribution of scores}
\label{sec:score_distribution}
We visualize the score distribution of several models in Figure~\ref{fig:score_distribution}. 
More figures of distributions are in Appendix~\textcolor{cvprblue}{O}. In generation tasks, the score densities of open-source models are heavily concentrated near 0.0-0.2, while those of GPT-Image-1 are concentrated near 0.8-1.0. On the contrary, in the editing tasks, all models see a similar trend, with score densities concentrated at the bottom. GPT-Image-1 shows slightly better performance than the best open-source model, OmniGen2, with the score densities at higher levels wider in this task. These observations are consistent with findings \circleone, \circletwo in Section~\ref{sec:findings}. For \textit{Vis}, the score densities concentrate at the top in the generation tasks, while in the editing tasks the scores of most models obviously see an upward shift compared to \textit{Cor}. This also supports the ``visually plausible but semantically mismatch'' findings in Section~\ref{sec:findings}.

\subsubsection{``Knowing how to plan'' v.s. visualization}
\label{sec:gap}
Figure~\ref{fig:gap_plan_gen} further reports the gap between textual planning and generation.
Detailed process and other cases are in Appendix~\textcolor{cvprblue}{P}. We discover that, 3 of 4 UMMs already know how to plan and give meaningful textual outputs in required JSON format, like ``plan steps:'' and ``operation->target''. Only Janus-4o fails to follow the instruction, and cannot show the textual plan. However, all four models fail to turn these language-level capabilities into real useful images. Consequently, it is still challenging for UMMs to let understanding really benefits generation in this domain.

\vspace{-1mm}
\subsection{Case and Failure Study}
\vspace{-1mm}
\label{sec:case_study}
\looseness=-1 Figure~\ref{fig:case} shows some cases (more in Appendix~\textcolor{cvprblue}{Q}). See generation sub-tasks, some open-source models (like Bagel, FLUX.1-dev) can generate the basic ``structure'' related to the query (like a curved line on a place with foods on the ground for route planning, or some boxes with texts for workflow diagramming), but they fail to generate queried details (like real food name, place, real steps of work, etc.) Other ones generate things more irrelevant, like Janus-4o here. In contrast, closed-source models make fewer mistakes. Gemini3-Pro-Image and GPT-Image-1 can generate generally correct images. Seeing editing sub-tasks, all models perform worse: mistakes like no editing (Step1X-Edit), image style totally changed (Gemini3-Pro-Image), and the route not being completed often happen. This study supports the findings in Section~\ref{sec:findings}, and underscores the need for improvements on layout understanding, fine-grained reasoning and so on for UMMs.

\section{Conclusion}
\label{sec:conclusion}
We present PlanViz, a benchmark for evaluating planning-oriented image generation and editing in computer-use tasks. We introduce diverse sub-tasks with high-quality human annotations and a specific evaluation metric, PlanScore. PlanViz reveals inconsistency on performance of current UMMs and exposes clear limitations of them, especially in planning-intensive image editing. Our results also indicate that satisfying constraints here remains a major challenge beyond surface-level image quality. Aspects like code rendering and the gap between textual planning and real generation are further explored. We hope that PlanViz will facilitate future research on bridging understanding and generation for real-world computer-use scenarios.
{
    \small
    \bibliographystyle{ieeenat_fullname}
    \bibliography{main}
}

\appendix


\end{document}